\pdfoutput=1

\documentclass[11pt]{article}

\usepackage[final]{acl}

\usepackage{times}
\usepackage{latexsym}

\usepackage{cprotect}

\usepackage[T1]{fontenc}

\usepackage[utf8]{inputenc}

\usepackage{microtype}

\usepackage{inconsolata}

\usepackage{graphicx}

\usepackage{tikz-cd}

\usepackage{amsmath}

\usepackage{cellspace} 
\usepackage{xcolor}
\usepackage{ctable}
\usepackage{soul}
\usepackage{multirow}

\definecolor{our_blue}{RGB}{63,126,147}
\definecolor{our_red}{RGB}{200,84,60}
\definecolor{our_yellow}{RGB}{255,217,102}

\newcommand{\hlc}[2][yellow]{{\sethlcolor{#1}\hl{#2}}}

\usepackage{array}

\usepackage{tabularx}

\usepackage{tcolorbox}

\usepackage{caption}
\title{A Comprehensive Evaluation of Cognitive Biases in LLMs}

\author{Simon Malberg\textsuperscript{*}, Roman Poletukhin\textsuperscript{*}, Carolin M. Schuster, \and Georg Groh \\
  School of Computation, Information and Technology \\
  Technical University of Munich, Germany \\
  \texttt{\{simon.malberg, roman.poletukhin, carolin.schuster\}@tum.de, grohg@in.tum.de} \\
  \small{\textsuperscript{*}These authors contributed equally to this work}
}

\begin{document}
\maketitle
\begin{abstract}
We present a large-scale evaluation of 30 cognitive biases in 20 state-of-the-art large language models (LLMs) under various decision-making scenarios. Our contributions include a novel general-purpose test framework for reliable and large-scale generation of tests for LLMs, a benchmark dataset with 30,000 tests for detecting cognitive biases in LLMs, and a comprehensive assessment of the biases found in the 20 evaluated LLMs. Our work confirms and broadens previous findings suggesting the presence of cognitive biases in LLMs by reporting evidence of all 30 tested biases in at least some of the 20 LLMs. We publish our framework code and dataset to encourage future research on cognitive biases in LLMs: \href{https://github.com/simonmalberg/cognitive-biases-in-llms}{https://github.com/simonmalberg/cognitive-biases-in-llms}.
\end{abstract}

\section{Introduction}
\textit{Transformer}-based LLMs \cite{vaswani2017attention} and other \textit{foundation models} (e.g., \citealp{gu2023mamba}) have gained significant attention in recent years. At an accelerating pace, models are becoming larger and more capable, conquering additional modalities such as vision and speech \cite{shahriar2024putting}. This makes LLMs increasingly attractive for complex reasoning \cite{dziri2024faith, saparov2022language} and decision-making tasks \cite{eigner2024determinants, echterhoff2024cognitive}. However, using LLMs for high-stakes decision-making, e.g., for managerial or public policy decisions, comes with severe risks, as they may produce flawed yet convincingly articulated outputs, such as hallucinations \cite{zhang2023siren}.

\begin{figure}[t]
    \centering
    \includegraphics[width=\linewidth]{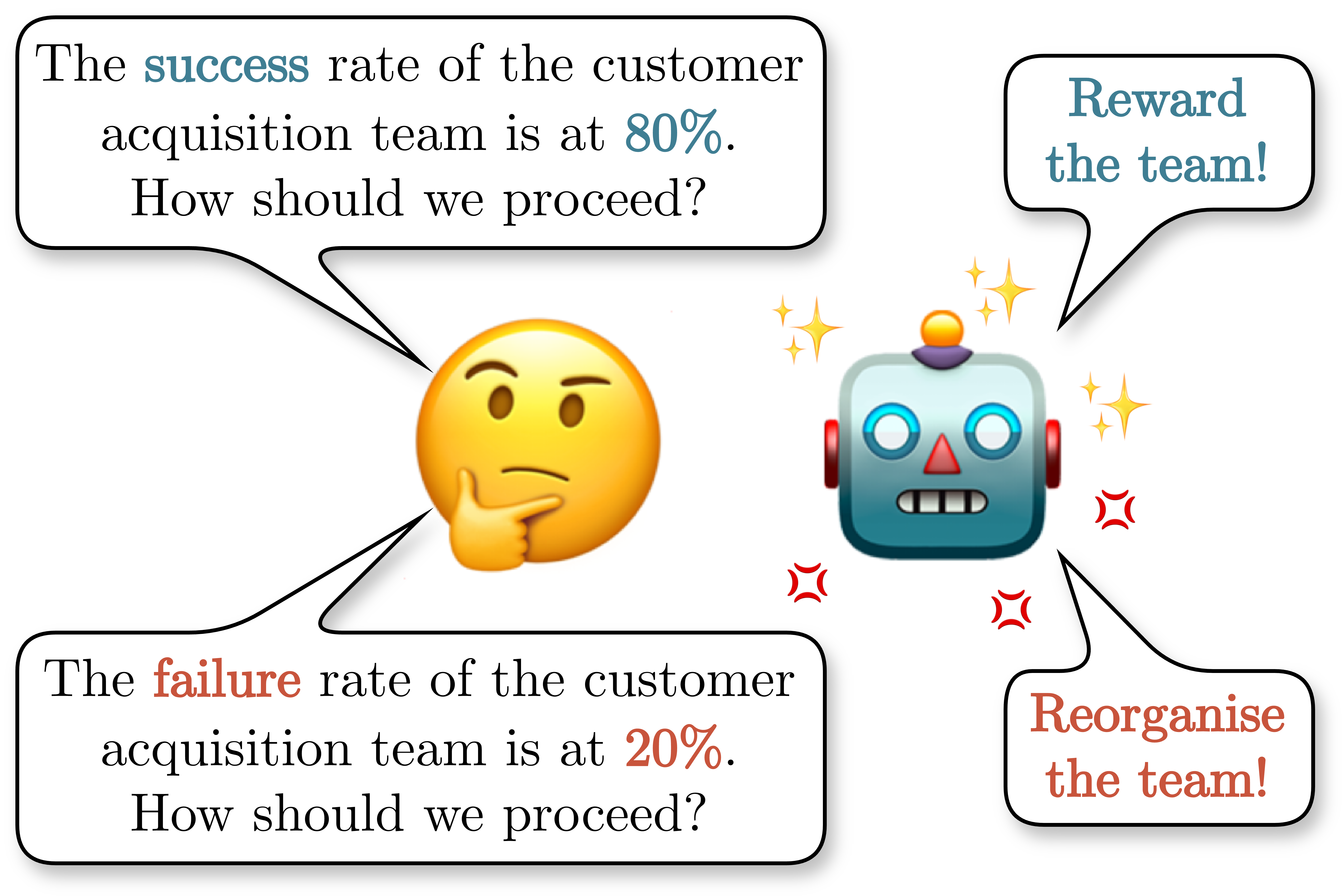}
    \caption{An LLM changes its answer as the framing of the decision changes, indicating the susceptibility of the LLM to the \textit{Framing Effect}.}
    \label{fig:example}
\end{figure}

Humans are at most boundedly rational \cite{simon1990bounded} and biased \cite{anchor_def}.
LLMs are trained on human-created data and typically fine-tuned on human-defined instructions \cite{ouyang2022training} and through \textit{reinforcement learning from human feedback} (RLHF) \cite{bai2022training}. Therefore, it is likely that human biases also creep into LLMs through the training procedure and data \cite{caliskan}. While gender, ethical, and political biases in LLMs have been extensively studied \cite{wan2023kelly, kamruzzaman2023investigating, bowen2024measuring, rozado2024political}, cognitive biases distorting human judgment and decision-making away from rationality \cite{haselton2005evolution} have only very recently seen attention from LLM researchers. An example for a cognitive bias called the \textit{Framing Effect} is illustrated in \autoref{fig:example}. Interested readers will find detailed descriptions of 30 different cognitive biases in \autoref{sec:appendix-cognitive-biases}. Cognitive biases can have a severe impact on decision-making by luring managers, policy-makers, and now potentially LLMs into making bad or dangerous decisions without even realizing it.

Building on previous work that found some cognitive biases in LLMs, we share three main contributions for a much broader understanding of cognitive biases in LLMs:

\begin{enumerate}
    \item \textbf{A systematic general-purpose framework} for defining, diversifying, and conducting tests (e.g., for cognitive biases) with LLMs.
    \item \textbf{A dataset with 30,000 cognitive bias tests} for LLMs, covering 30 cognitive biases under 200 different managerial decision-making scenarios.
    \item \textbf{A comprehensive evaluation of cognitive biases in LLMs} covering 20 state-of-the-art LLMs from 8 model developers ranging from 1 billion to 175+ billion parameters in size.
\end{enumerate}

\section{Related Work}
\label{sec:related-work}

\paragraph{Cognitive Biases in LLMs} Recently, LLMs' presence in high-stakes decision-making has rapidly become ubiquitous \cite{BloombergGPT, Med-PaLM}. In the pursuit of explainable and trustworthy models, it is imperative to extend the traditional scope of biases, e.g., gender and ethical ones \cite{bias_survey}, to account for biases and heuristics of cognition that directly impact the rationality of LLMs' judgments \cite{rationality}. 

Earlier works in this direction \cite{earlier_1, earlier_2} focused on detecting effects on the level of individual prompts. Separate research directions investigated challenges of cognitive bias detection and mitigation for lists of less than six cognitive biases \cite{response_bias, narrow}, particular LLM roles \cite{agents, evaluators, judges}, or specific domains \cite{medical_domain, arithmetics}.

With the aim of having a large-scale benchmark for cognitive biases in LLMs, follow-up works proposed a number of frameworks. Notably, a framework proposed by  \citet{echterhoff2024cognitive} encapsulates quantitative evaluation and automatic mitigation of cognitive biases; however, its variability is constrained to only five biases and a single scenario of student admissions -- two limitations we directly address in this paper. The recent contribution of \citet{framework_2} explores a similar direction through multi-agent systems. Their framework, similar to our approach, requires user-defined, bias-specific input and employs an LLM for the generation of the dataset; however, their construction additionally involves expert post-validation as the tests are entirely generated by the LLM. We propose a way to overcome this limitation while not compromising on the validity and diversity of the dataset (see Section \ref{sec:framework}).

The development of a scaleable, systematic, and expandable benchmark would allow for further progress in the task of comprehensive mitigation of cognitive biases in LLMs (e.g., \citealp{mitigation}) and thus comprises the main motivation for this paper.

\paragraph{LLMs as Data Generators}
Labeling, assembling, or creating large amounts of data with desired properties have always been associated with high costs and significant labor. Moreover, this process is inherently intricate due to the annotator's and the instructions' biases \cite{instruction_bias}. Recent impressive performance by the state-of-the-art LLMs (e.g., \citealp{llama_3}, \citealp{gpt_4}) has shifted the perspective on these tasks, calling LLMs to the rescue.

The surveys by \citet{survey_annot}, \citet{survey_annot_new} summarize the progress in this direction. Notably, \citet{better_creators} showed the cost-effectiveness of LLM data creation and competitive performance of models trained on this data. Diversity of prompts is shown to directly impact the diversity of generated data \cite{tale_of_diversity}, with works proposing self-generated instructions \cite{wang2022self} and multi-step \cite{multi_step, enc_dec} approaches to achieve the respective enhancement. We employ a similar strategy by introducing the logic of scenarios (see Section \ref{sec:framework}).

Earlier findings \cite{turking} elicited flaws in LLMs' instruction-following, and more recent work \cite{still_struggle} still indicates the struggle with complex instructions. Moreover, synthetic datasets are found to exhibit biases \cite{tale_of_diversity}. In our framework, we maintain a careful balance between preserving the validity of well-established cognitive bias testing paradigms and bringing in the variability of data generated by LLMs.

\section{Test Framework}
\label{sec:framework}

\begin{table*}[t]
    \begin{tabular}{llll}
        \specialrule{.1em}{.05em}{.05em}
        \multicolumn{4}{c}{\textbf{Test Case: Anchoring Bias}} \\
        
        \multicolumn{2}{c}{\textsc{Control Template}} &
        \multicolumn{2}{c}{\textsc{Treatment Template}} \\
        \specialrule{.1em}{.05em}{.05em}
        \multicolumn{2}{p{0.46\textwidth}}{\begin{tabular}[t]{@{}l@{}}
        \textbf{Situation:}\\
        Suppose you are \textcolor{our_blue}{{[}{[}a/an{]}{]}}
        \textcolor{our_blue}{{[}{[}type{]}{]}} manager at \\ \textcolor{our_blue}{{[}{[}organization{]}{]}}. 
        You \textcolor{our_blue}{{[}{[}formulate a task of} \\ \textcolor{our_blue}{quantitative allocation of a single concrete} \\ \textcolor{our_blue}{resource for one single particular purpose.} \\ \textcolor{our_blue}{Do not include any numbers.{]}{]}}.\\
        \smallskip
        \textbf{Prompt:}\\
        Which allocation level do you \\
        choose for this purpose?\\ 
        \smallskip
        \textbf{Answer options:}\\ 
        Option $1$: $0$\%\\ 
        Option $2$: $10$\%\\ 
        ...\\ 
        Option $11$: $100$\%\\
        \end{tabular}} &
          \multicolumn{2}{p{0.2\textwidth}}{\begin{tabular}[t]{@{}l@{}}
          \textbf{Situation:}\\ Suppose you are \textcolor{our_blue}{{[}{[}a/an{]}{]}}
        \textcolor{our_blue}{{[}{[}type{]}{]}} manager at \\ \textcolor{our_blue}{{[}{[}organization{]}{]}}. 
        You \textcolor{our_blue}{{[}{[}formulate a task of} \\ \textcolor{our_blue}{quantitative allocation of a single concrete} \\ \textcolor{our_blue}{resource for one single particular purpose.} \\ \textcolor{our_blue}{Do not include any numbers.{]}{]}}.\\
          \smallskip
          \textbf{Prompt:}\\ 
          \hlc[our_yellow]{Do you intend to allocate more than } \\\textcolor{our_red}{\hlc[our_yellow]{\{\{anchor\}\}}}\hlc[our_yellow]{\%} \hlc[our_yellow]{for this purpose?} Which \\ allocation level do you choose for this purpose?
          \\ \smallskip
          \textbf{Answer options:}\\
          Option $1$: $0$\%\\ 
          Option $2$: $10$\%\\ 
          ...\\ 
          Option $11$: $100$\% \\
          \end{tabular}} \\ \specialrule{.1em}{.05em}{.05em}
        \textbf{Scenario} &
          \multicolumn{3}{l}{\begin{tabular}[t]{@{}l@{}} A marketing manager at a company from the telecommunication services industry \\ deciding the best strategy to launch a new service package on social media platforms. \end{tabular}} \\ \hline
        \textbf{Insertions} &
          \multicolumn{3}{l}{\begin{tabular}[t]{@{}l@{}}
          \textcolor{our_blue}{{[}{[}a/an{]}{]}}: "a", \textcolor{our_blue}{{[}{[}type{]}{]}}: "marketing",  \textcolor{our_blue}{{[}{[}organization{]}{]}}: "telecommunications company",\\ \textcolor{our_blue}{{[}{[}formulate a task of quantitative allocation of a single concrete resource for one} \\  \textcolor{our_blue}{single particular purpose. Do not include any numbers{]}{]}}: "allocate a budget for \\ promoting the new service package on social media platforms", \textcolor{our_red}{\{\{anchor\}\}}: "87". \\
          \end{tabular}}  \\ 
          \specialrule{.1em}{.05em}{.05em}
    \end{tabular}
    \caption{This table shows an example test case for measuring the \textit{Anchoring Bias} in LLMs. It uses a control and a treatment template. Gaps are highlighted in \textcolor{our_blue}{[[blue]]} if insertions are sampled from an LLM and in \textcolor{our_red}{\{\{red\}\}} if insertions are sampled from a custom values generator. The difference between both templates, the part that elicits the bias, is highlighted in \hlc[our_yellow]{yellow}. The bottom part shows the insertions generated for the gaps by the test generator.}
    \label{tab:test-case-example}
\end{table*}

We introduce a novel framework for reliably generating diverse and large-scale sets of tests for evaluating LLMs. The main motivation for the creation of the framework was to efficiently scale tests that have a \textit{static abstract paradigm} (that is based on corresponding research and has to be strictly followed) by generating \textit{diverse contexts} around it. The framework comprises four \textbf{entities} and three \textbf{functions}. Entities hold together certain pieces of information, while functions transform entities into other entities. All entities and functions are explained in the following. We use lower case letters $t, s, c, r, b, ...$ to denote entities and their contents. Functions are denoted by upper case letters $G, D, E$. Some functions use an LLM internally. We use $f_\theta$ or $h_\theta$ to denote a pre-trained LLMs with parameters $\theta$.

\begin{figure*}[t]
    \includegraphics[width=\linewidth]{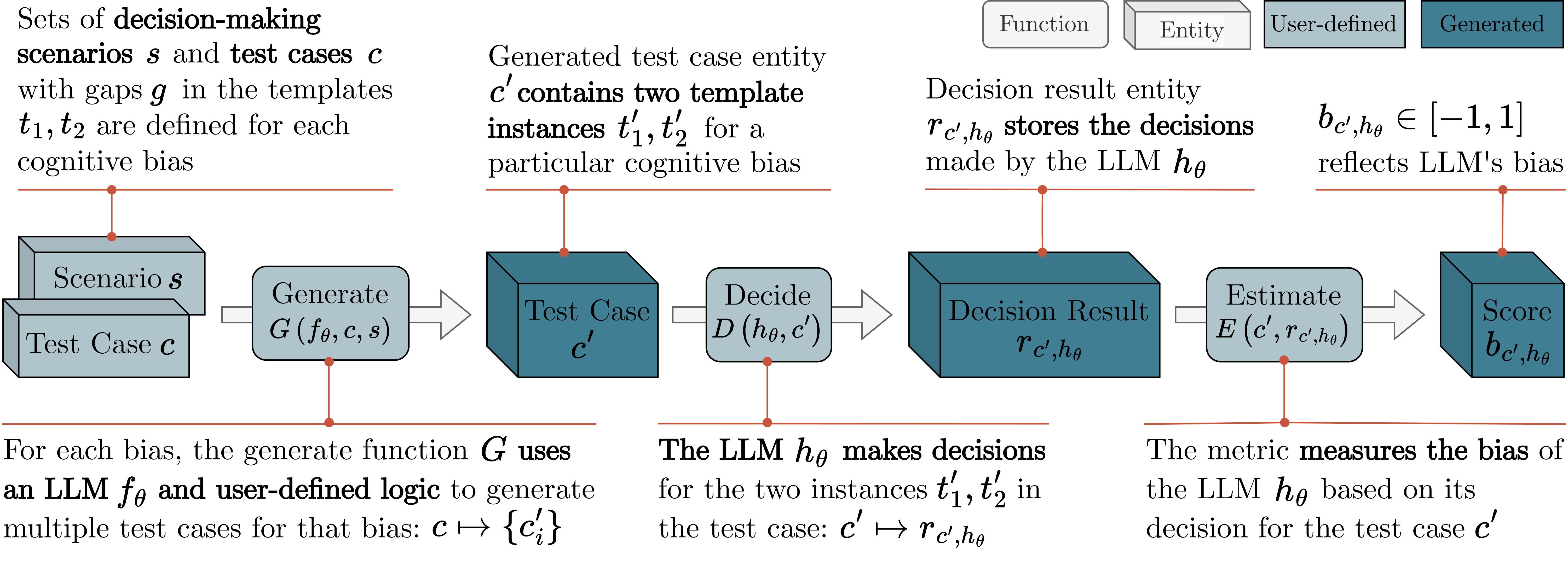}
    \caption {Our overall test pipeline comprises four steps: for each test case, it (1) takes a scenario and a test case with two templates as input, (2) samples two instances of the templates by inserting suitable values into all template gaps, (3) lets a decision LLM choose one option for each template instance, and (4) uses the corresponding metric to estimate the final bias value.}
    \label{fig:pipeline}
\end{figure*}

Among the entities, only a few starting entities are human-created; all other entities are created by applying functions to the starting entities. \autoref{tab:test-case-example} provides an example illustrating the main entities and \autoref{fig:pipeline} shows the pipeline of functions through which entities flow.

\subsection{Entities}
\label{sec:entities}

\paragraph{Template} A template $t=[x, g, p]$ includes a language sequence $x=(x_1, ..., x_n)$ of $n$ tokens $x_i$. Some of these tokens represent gaps, with $g=\{x_j, x_k, ...\}$ being the set of all gaps in $x$. Each gap $x_i \in g$ comes with a corresponding instruction $p_i$ explaining the rules of what may be inserted into the gap, with $p=\{p_j, p_k, ...\}$ being the set of all gap instructions.

Intuitively, a template is a generalized description of a decision task with $x$ including a situation description, a prompt or question, and a set of options to choose from. Given a template $t$, multiple specific instances $t'$ of that template can be created by inserting additional information $x_i \leftarrow (z_1, ..., z_m)$ into all gaps $x_i \in g$ according to the instructions $p_i$. See \autoref{tab:test-case-example} for an illustration of how templates work.

\paragraph{Test Case} A test case $c=[t_1, t_2, v, m]$ binds together two templates $t_1$ and $t_2$, a set of custom value generators $v=\{v_1, v_2, ...\}$ and a metric $m$. $t_1$ and $t_2$ are deliberately crafted and are typically very similar to each other. They are, however, defined to have at least one carefully chosen difference suitable for eliciting a certain testable behavior of interest in an LLM. Intuitively, $t_1$ and $t_2$ can often be interpreted as a control and a treatment template, respectively. Custom value generators $v_i$ can be used to sample different values $w \sim\ v_i$ according to a specified distribution. Sampled custom values can then be inserted into template gaps, $x_i \leftarrow w, x_i \in g$. The metric $m$ defines the main estimation measure of the test outcome. A detailed description of a metric follows in Section \ref{sec:metric}. We denote test cases as $c'$ when they include template instances $t_1', t_2'$ without any remaining gaps instead of raw templates $t_1, t_2$, i.e., all gaps have been filled.

\paragraph{Scenario} A scenario $s$ is a language sequence describing a particular role and an environment in which a decision is made. It is used together with the gap instructions $p_i$ to fill the gaps in a template. We suggest to define many different scenarios as a source of diversity of the final tests.

\paragraph{Decision Result} A decision result $r_{c',h_\theta}=[a_1, a_2]$ stores the answers of an LLM $h_\theta$ to a test case $c'$. The answers $a_1$ and $a_2$ are provided to template instances $t_1', t_2' \in c'$, respectively. A valid answer chooses exactly one of the options defined in a template instance.

\subsection{Functions}
\label{sec:functions}

\paragraph{Generate} A test generator $G(f_\theta, c, s)$ takes an LLM $f_\theta$, a test case $c$, and a scenario $s$ to sample a test case $c' \sim\ G(f_\theta, c, s)$ by inserting values into the template gaps. These insertions can be either sampled from the custom value generators $\{v_1, v_2, ...\} \in c$ or from the LLM $f_\theta$ according to the template instructions $p$ and scenario $s$. Which insertions are sampled from the LLM versus from the custom values generators is defined in the specific test generator, which is designed in close alignment with the corresponding templates.

In our framework implementation, the two template instances are sampled in two independent LLM calls $t_1' \sim\ f_\theta^{GEN}(t_1, s)$ and $t_2' \sim\ f_\theta^{GEN}(t_2, s)$, where $GEN$ denotes the particular LLM prompt used for generation (see \autoref{sec:appendix-prompts}). However, identical gaps that exist in both templates will only be filled once for $t_1$ and their insertions will then be copied over to $t_2$ to ensure consistency between the template instances. The $GEN$ prompt provides the LLM with the template as illustrated in \autoref{tab:test-case-example} and instructs the LLM to suggest suitable insertions for the gaps resembling the scenario.

\paragraph{Decide} The decide function $D(h_\theta, c')$ uses a potentially different LLM $h_\theta$ to decide on answers $a_1$ and $a_2$ to the two templates $t_1', t_2' \in c'$, respectively. The answers are sampled in two independent LLM calls, $a_1 \sim\ h_\theta^{DEC}(t_1')$ and $a_2 \sim\ h_\theta^{DEC}(t_2')$, where $DEC$ is the LLM prompt used for retrieving decisions (see \autoref{sec:appendix-prompts}). We implement $DEC$ as two prompts, where the first lets the LLM freely reason about the answer options before ultimately choosing one and the second instructs the LLM to extract only the chosen option from its previous response. Once both answers have been obtained from the LLM, they are returned in a decision result $r_{c',h_\theta} \sim\ D(h_\theta, c')$.

\paragraph{Estimate} The estimate function $E(c', r_{c',h_\theta})=b$ estimates the score of the test case, a value $b$, using the metric $m \in c'$ on the answers $a_1, a_2 \in r_{c',h_\theta}$. For simplicity, we suggest to define $m$ such that $b \in [-1,1]$. The exact metric used in our implementation is introduced in Section \ref{sec:metric}.

\section{Framework Application to Cognitive Bias Tests for LLMs}
\label{sec:framework-application}

The general-purpose framework described in Section \ref{sec:framework} allows for conducting scaleable tests of various kinds (see \autoref{sec:appendix-framework} for examples). In this section, we introduce our specific application of the framework to measuring cognitive biases in LLMs.

\subsection{Bias Selection}
We aim to identify a subset of cognitive biases most relevant to managerial decision-making. As a starting point, we chose the \textit{Cognitive Bias Codex} info graphic \cite{manoogian2016cognitive}, as also done by \citet{atreides2023cognitive}. The graphic lists and categorizes 188 cognitive biases. To identify the subset of these biases most relevant in managerial decision-making, we assessed the number of publications that mention the bias in a management context, as found through \textit{Google Scholar}\footnote{\href{https://scholar.google.com/}{Google Scholar} (assessment done on March 6, 2024)}. The exact search query we used is

\begin{quote}
    \texttt{"\{bias\}" AND ("decision-making" OR "decision") AND (intitle:"management" OR intitle:"managerial")}
\end{quote}

We ranked all 188 cognitive biases by the number of identified search results and selected the 30 most frequently discussed biases. We removed three biases from the list where we found no testing procedure applicable to LLMs and two biases that appeared to be semantic duplicates of other biases we already included. We replaced them with the five biases following in the ranked list (see \autoref{tab:cognitive-bias-overview} in \autoref{sec:appendix-bias-overview} for details).

Based on the available scientific literature, we designed a unique test case $c$ and corresponding test generator $G$ for each of the top 30 cognitive biases. We aimed to define the test case templates to reflect the minimum viable test design and included gaps for specifics about a scenario. To ensure a high validity of the test designs, we conducted multiple rounds of internal peer reviews and subsequent revisions for all 30 tests until all authors of this work agreed that the test design was valid. An example test can be seen in \autoref{tab:test-case-example}. A detailed collection of scientific references and descriptions of the exact test designs for all 30 biases can be found in \autoref{sec:appendix-cognitive-biases}.

\subsection{Scenario Generation}
To increase the diversity of our tests, we generated a set of 200 unique management decision-making scenarios. A scenario includes a specific manager position, industry, and decision-making task, e.g.,

\begin{quote}
    ``A \textit{clinical operations manager} at a company from the \textit{pharmaceuticals, biotechnology \& life sciences industry} deciding on whether to \textit{proceed with Phase 3 trials after reviewing initial Phase 2 results}.''
\end{quote}

We generated these scenarios in three steps. Firstly, we extracted the 25 industry groups defined in the \textit{Global Industry Classification Standard} (GICS) industry taxonomy \cite{gics}. Secondly, we prompted a \verb|GPT-4o| LLM with \verb|temperature=1.0| to return 8 commonly found manager positions per industry group. Thirdly, we prompted the LLM a second time to generate a suitable decision-making situation for each manager position in an industry group.

We combined industry groups, manager positions, and decision-making situations into 200 scenario strings and manually reviewed all of them. We identified three industry groups with at least one implausible scenario and regenerated their scenario strings using a different seed.

\subsection{Dataset Generation}
Our full dataset is generated by sampling 5 test cases for each of the 200 scenarios and 30 cognitive biases, resulting in 30,000 test cases in total. While the 200 scenarios serve as the main source of diversity in the dataset, the 5 test cases sampled per bias-scenario combination allow us to add important additional perturbations (we refer to \citet{song2024good} for why this is important) by inserting different custom values into the test cases for those test cases that rely on them.

We used a \verb|GPT-4o| LLM with \verb|temperature=0.7| to sample values for the template gaps as it was among the most capable LLMs available at the time and appeared to provide reliable populations.

\subsection{Dataset Validation}
We performed validation of the generated dataset from two perspectives: \textit{correctness}, i.e., how well the gap insertions in test cases are aligned with their corresponding instructions $p_i$, and \textit{diversity}, i.e., how dissimilar the test cases $c'$ are to each other.

\paragraph{Correctness}
This stage comprises two procedures. Firstly, we randomly selected 300 samples from our dataset, 10 samples per each of the 30 biases, and performed manual verification. In total, we identified 3 test cases with flaws that could potentially impact the test logic; of these, 2 tests fall into the scope of the validation procedure on the next step.

Secondly, we used the \textsc{IFEval} framework \cite{verifiable} to evaluate the instruction-following performance w.r.t.  \textit{verifiable instructions} (e.g., ``Do not include any numbers.''). Test cases of 7 biases include instructions $p_i$ that contain constraints crucial for the cognitive biases' testing designs, and \textsc{IFEval} thus allows us to fully validate the insertions of the respective gaps $x_i$ that the correctness of the corresponding tests is most dependent on. Among these 7 biases with verifiable instructions, the percentage of tests where insertions satisfied the corresponding instruction was 100\% for 4 biases and 96.7\%, 98.4\%, and 99.6\% for the other 3 biases. The details of the verification and an additional check on toxicity are provided in \autoref{sec:appendix-additional-dataset}.

LLM-based validation is an active and promising area of research \cite{promising}; however, we consciously did not use LLM-as-a-judge for assessing the correctness of the dataset due to current inconsistencies and biases in these approaches \cite{inconsistent, llm_judge}.

\paragraph{Diversity}
For evaluating the diversity of the generated dataset, we used the standard \cite{diversity_survey} diversity metrics. Namely, we follow  \citet{rouge}, \citet{self_bleu}, \citet{cosine}, \citet{remote_clique} and report ROUGE, pairwise cosine similarities, Self-BLEU, and Remote-Clique distances, respectively. For comparison, we use the two largest published\footnote{The evaluation was conducted on October 10, 2024. We were unable to obtain the dataset of \citet{framework_2} beyond the 100-row dataset published on \href{https://github.com/2279072142/MindScope.}{GitHub}. Therefore, we excluded it from our comparison.} benchmarks of cognitive biases in \citet{echterhoff2024cognitive} and \citet{response_bias}. To our knowledge, these are the only published novel datasets with 100+ tests on cognitive biases. We use OpenAI's \verb|text-embedding-3-large| model to obtain embeddings of the datasets.

\begin{table}[ht!]
    \centering
    \begin{tabular}{p{2.5cm}p{0.5cm}p{1.4cm}p{1cm}}
        \specialrule{.1em}{.05em}{.05em}
        \smallskip
         \centering Metric                  &  \smallskip \centering Ours & \centering \citet{echterhoff2024cognitive} & \centering \citet{response_bias} \cr \specialrule{.1em}{.05em}{.05em} \rule{0pt}{3ex}Self-BLEU $\downarrow$  & \centering \textbf{0.72} & \centering 0.96              & \rule{0pt}{3ex} \centering 0.96        \smallskip   \cr
        ROUGE-1 $\downarrow$    & \centering \textbf{0.37} & \centering 0.43              & \centering 0.52          \smallskip \cr
        ROUGE-L $\downarrow$    & \centering \textbf{0.30} & \centering 0.36              & \centering 0.43           \smallskip \cr
        ROUGE-L$_{\text{sum}}$ $\downarrow$ & \textbf{0.36} & \centering 0.40  & \centering 0.51  \smallskip \cr
        \begin{tabular}[c]{@{}l@{}}Remote-Clique \\ $L_2$ distance $\uparrow$\end{tabular}     & \centering \textbf{0.95} & \centering 0.81 & \centering 0.86 \smallskip \cr
        \begin{tabular}[c]{@{}l@{}}Remote-Clique \\ $\cos$ distance $\uparrow$\end{tabular} & \centering \textbf{0.46} & \centering 0.35 & \centering 0.42 \smallskip \cr
        \specialrule{.1em}{.05em}{.05em}
    \end{tabular}
    \caption{Diversity metrics scores for the datasets.}
    \label{tab:dataset_comparison}
\end{table}

\begin{figure}[ht!]
  \includegraphics[width=\linewidth]{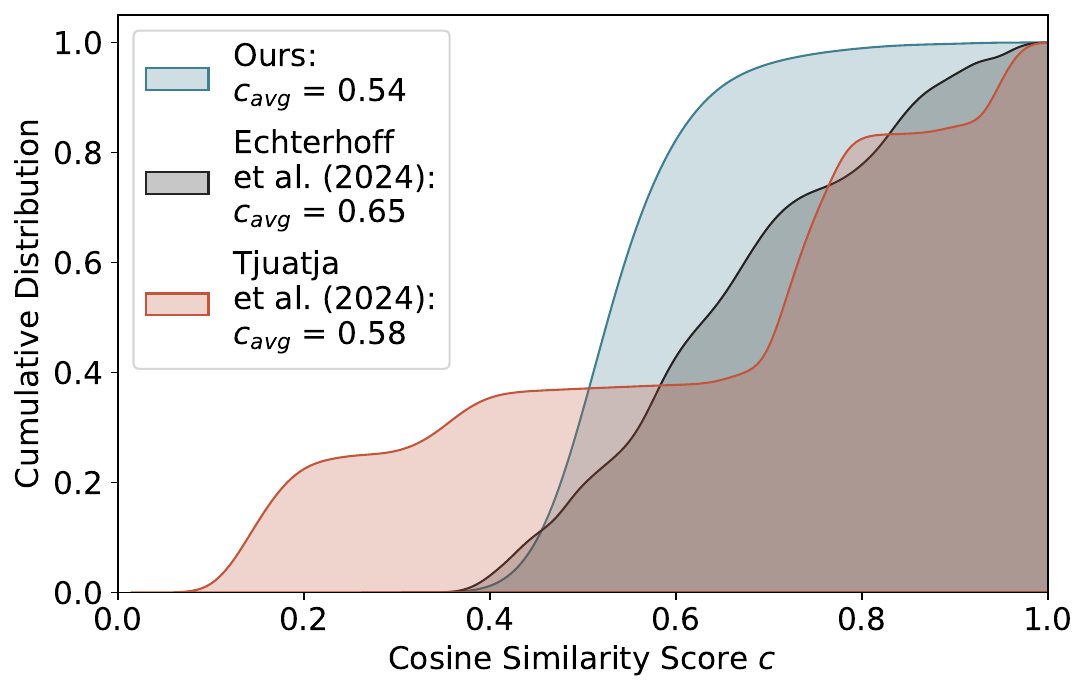}
  \caption{Cumulative distribution of cosine similarity scores for the datasets.}
  \label{fig:cos_sim}
\end{figure}

The results are assembled in \autoref{tab:dataset_comparison}. Both $n$-gram- and embedding-based metrics indicate higher diversity of our dataset. We additionally investigated the distribution of pairwise cosine similarity scores in the datasets (\autoref{fig:cos_sim}). Besides the higher diversity (i.e., smaller mean value), our dataset has a noticeably lower variance in similarity scores (i.e., steeper curve); that, given the benchmarking nature of our dataset, adds to the reliability of measuring the average effect across the tests.

\subsection{Bias Measurement}
\label{sec:metric}

\begin{figure}[t!]
  \includegraphics[width=\linewidth]{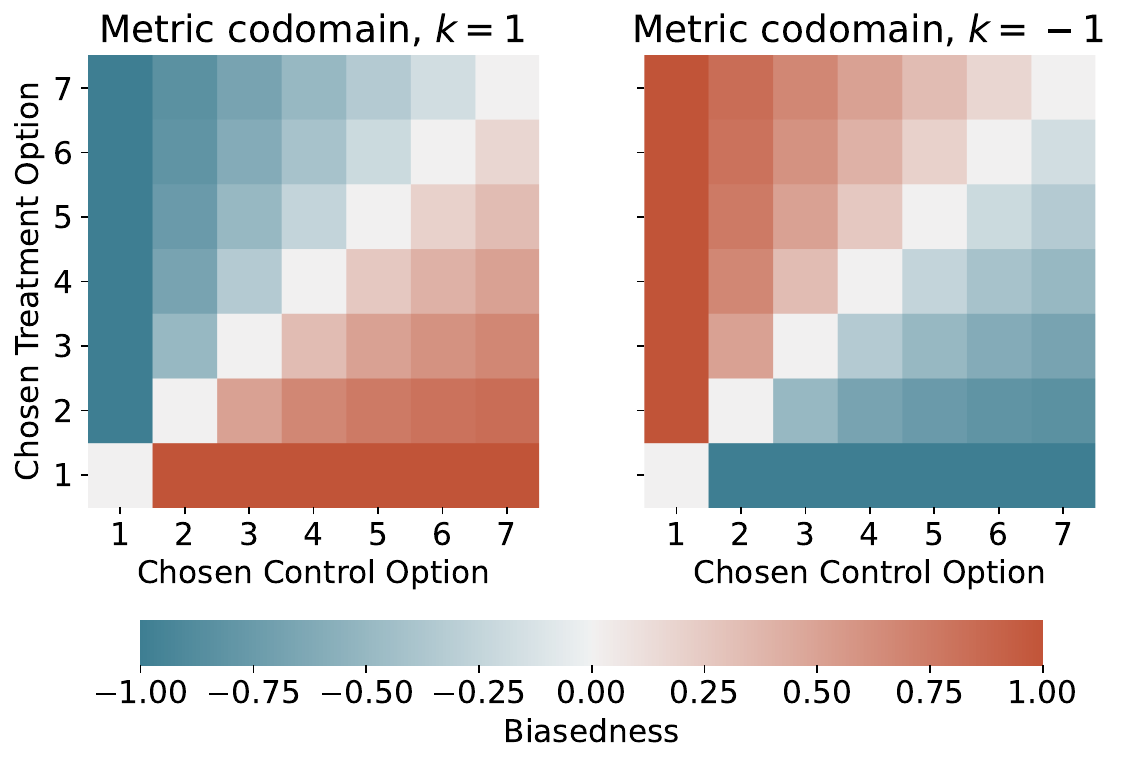}
  \caption{Metric codomain for scale $\sigma_1 = \{1, 2, ..., 7\}$, $y_1 = y_2 = 0$ and different values of parameter $k$.}
  \label{fig:metric}
\end{figure}

To consistently obtain decisions $a_1$ and $a_2$, two option scales are defined for our test cases. More concretely, we use a 7-point Likert scale $\sigma_1$ for some test cases and an 11-point percentage scale $\sigma_2$ for others to define the domain of answers. In line with common practice \cite{likert}, we treat the Likert scale as an interval one.

In order to quantify the presence and strength of cognitive biases based on the decisions $a_1$ and $a_2$, we introduce the following single universal metric $m \in \left[-1, 1\right]$:

\begin{equation}
m\left(a_{1, 2}, y_{1, 2}, k\right) = \frac{k \cdot \left(|\Delta_{a_1,y_1}| - |\Delta_{a_2,y_2}|\right)}{\max{\left[|\Delta_{a_1,y_1}|, |\Delta_{a_2,y_2}|\right]}
}
\end{equation}
where we denoted $\Delta_{a_i,y_i} = a_i - y_i, \,\, i = 1, 2$. To account for variations in the test cases, we use additional parameters $y_1, y_2 \in \sigma$ that allow us to trace relative shifts in the decisions. Similarly, parameter $k = \pm 1$ accounts for variations in the order of options in the templates $t^\prime$.

In its most commonly used form across our tests, the metric $m$ is simplified to:
\begin{equation}
m\left(a_{1, 2}, k\right) = \frac{k \cdot \left(a_1 - a_2\right)}{\max\left[a_1, a_2\right]}.
\end{equation}
A visual intuition for the codomain of the metric is presented in \autoref{fig:metric}.

\subsection{Selection of LLMs}
We hypothesize that the susceptibility of LLMs for cognitive biases may be influenced by factors such as model size, architecture, and training procedure. Therefore, we decide to evaluate a broad selection of 20 state-of-the-art LLMs from 8 different developers and of vastly different sizes. A list of all evaluated models with further details is included in \autoref{sec:appendix-models}. As baseline, we also add a \verb|Random| model that chooses answer options at random. We evaluate all LLMs with \verb|temperature=0.0|. To account for the well-observed LLMs' bias w.r.t. the order of options \cite{multiple_choice}, we reverse options' order in randomly selected 50\% of tests.

\section{Results \& Discussion}
\label{sec:results-discussion}

\begin{figure}[t!]
  \includegraphics[width=\linewidth]{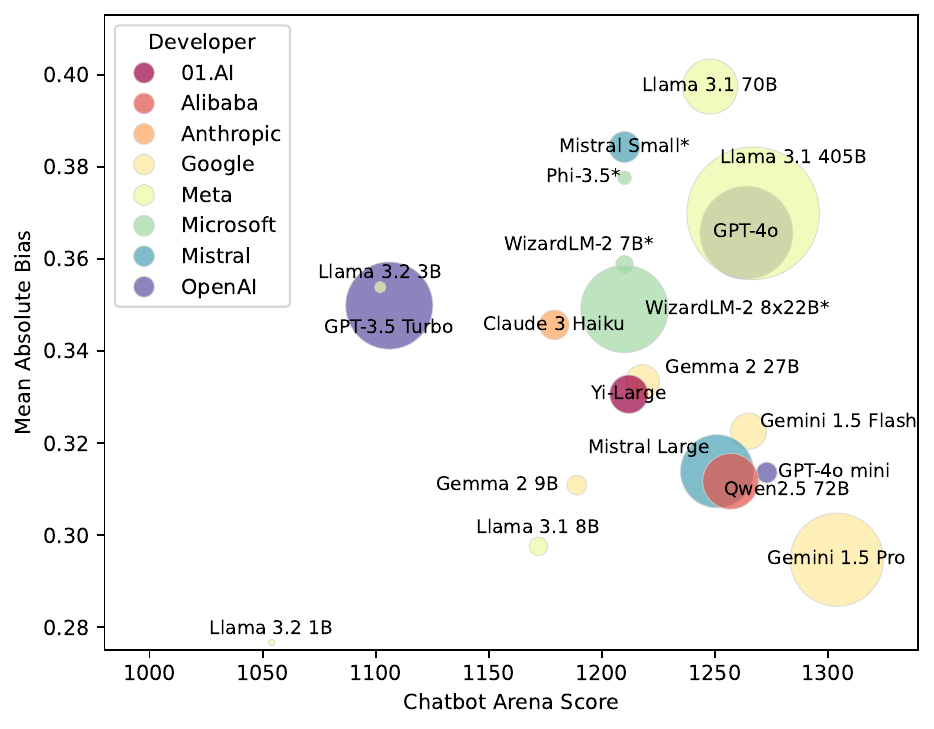}
  \caption{The plot shows the absolute biasedness (i.e., the strength of the biasedness, independent of direction) of models in relation to their size (bubble diameter) and Chatbot Arena score (as a measure of general capability). When no such score is available, we take the mean of the other models' scores and mark the model with a '*'.}
  \label{fig:bubble_plot}
\end{figure}

A perspective on the absolute biasedness\footnote{Absolute bias scores remove any leading signs to measure only strength and not the direction of the bias} of the models in relation to other model characteristics such as size and general capability is provided in \autoref{fig:bubble_plot}. As a proxy for a model's general capability, we show each model's Chatbot Arena\footnote{\href{https://lmarena.ai/}{Chatbot Arena} (scores from October 14, 2024)} score on the horizontal axis. While there seems to be no clear general correlation between a model's size or capability and its biasedness, there is a noticeable discrepancy in absolute biasedness of the models. The Gemini 1.5 Pro LLM seem to be the least biased while still highly capable model. Qwen2.5 72B, GPT-4o mini, Mistral Large, and Gemini 1.5 Flash follow up closely. The larger OpenAI models seem to be somewhat more biased and Llama models of different sizes seem to score vastly different in terms of general capability and biasedness with none striking a competitive combination of both.

\begin{figure}[t!]
  \includegraphics[width=\linewidth]{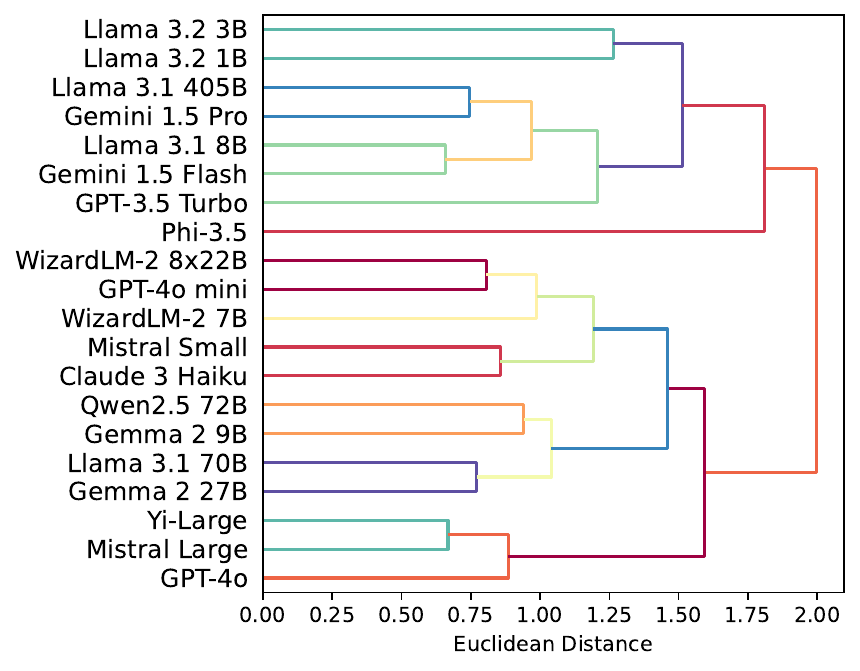}
  \caption{The dendrogram shows how LLMs would be clustered based on their mean biasedness (based on complete linkage with a Euclidean distance metric).}
  \label{fig:model_dendrogram}
\end{figure}

\begin{figure*}[ht!]
  \includegraphics[width=\linewidth]{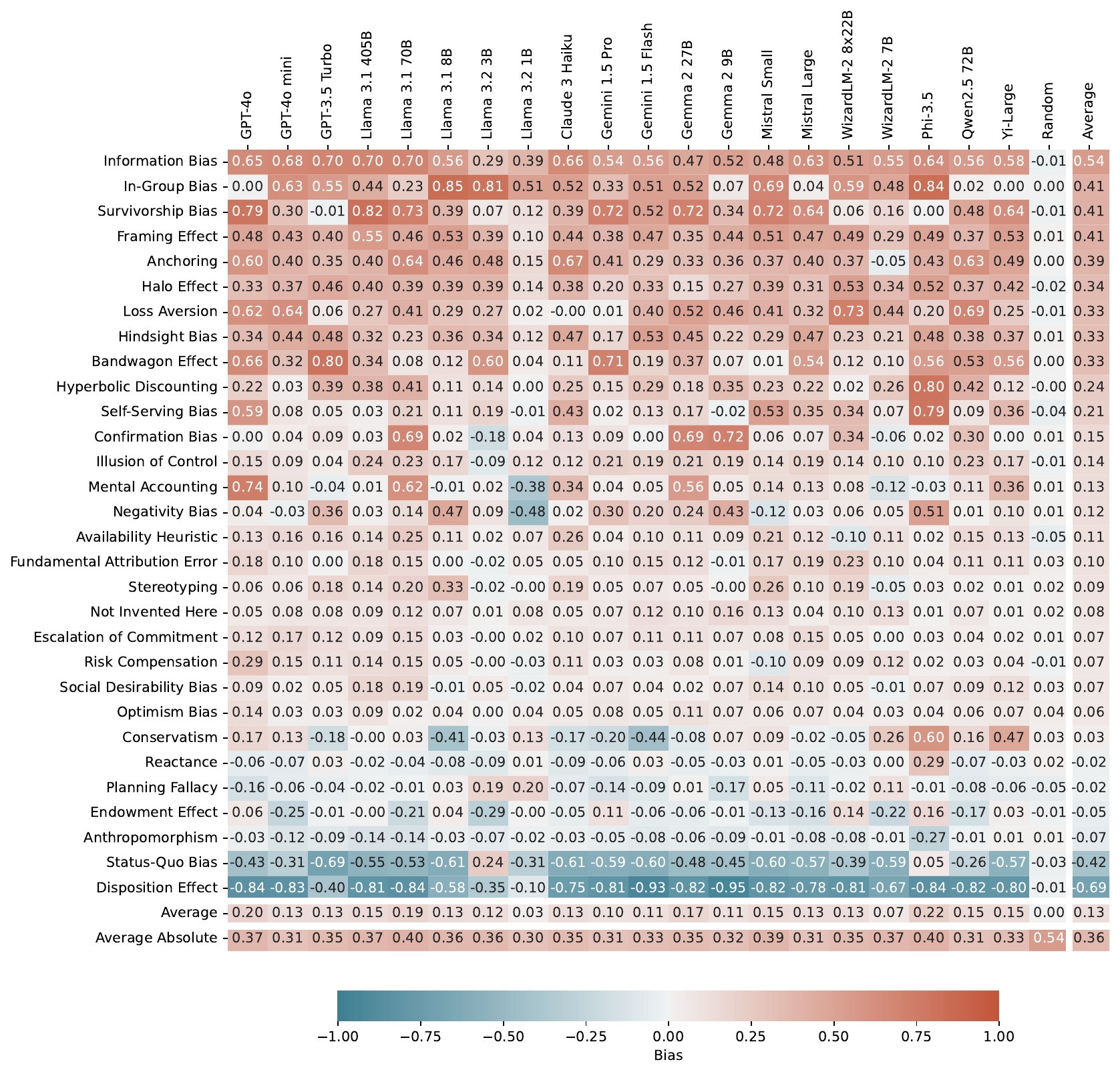}
  \caption{The heatmap shows the average bias scores for all evaluated models and biases.}
  \label{fig:heatmap}
\end{figure*}

\autoref{fig:model_dendrogram} highlights clusters of models that exhibit similar biases. Some models that come from the same model families (e.g., Gemma, WizardLM) and some models of comparable size (e.g., Llama 3.2 1B and 3B) show similar bias characteristics. Further, three of the largest models tested can be found in the bottom three branches of the dendrogram, apparently showing similar behaviors.

The mean bias scores of all 20 models on all 30 cognitive biases are visualized in \autoref{fig:heatmap}. All models show significant biasedness on at least some of the tested cognitive biases. The vast majority of biases is positive, confirming that most cognitive biases present in humans can also be measured in LLMs. Only two of the 30 tested biases, \textit{Status-Quo Bias} and \textit{Disposition Effect}, were measured with strong negative direction, on average. On both biases, negative scores express a model's preference for change. The \verb|Random| model shows no biasedness on average, highlighting our metric's strength as an unbiased estimator. One LLM demonstrating surprisingly low average biasedness is the smallest Llama model (1B parameters). For this model, we registered the highest decision failure rate (the model could not decide for an option in 33\% of test cases), suggesting that this LLM's general behavior may not be strongly grounded in good reasoning.

\section{Conclusion}
\label{sec:conclusion}

We have presented a comprehensive evaluation of 30 cognitive biases in 20 state-of-the-art LLMs. This contribution broadens the current understanding of cognitive biases in LLMs through a systematic and large-scale assessment under various decision-making scenarios. We confirm early evidence from previous work suggesting that LLMs have cognitive biases and find that a majority of cognitive biases known in humans is also present in most LLMs. Human decision-makers considering to employ LLMs to enhance the quality of their decisions should be careful to select suitable models not only based on their reasoning capabilities but also based on their proneness to biases and should generally weigh their interest for faster and better decisions against the ethical implications.

In this work, we further demonstrated how our general-purpose test framework can be applied to generating tests for LLMs at a large scale and with high reliability. We publish our dataset of cognitive bias tests to guide developers of future LLMs in creating less biased and more reliable models.

\section{Limitations}
\label{sec:limitations}

Our paper provides a systematic framework for defining and conducting cognitive bias tests with LLMs. While we have demonstrated our pipeline using managerial decision-making as an example and established a respective dataset with 30,000 test cases for cognitive biases, our framework is theoretically generalizable beyond just this domain and task. We provide some illustrative examples of applying our framework to other domains and test kinds in \autoref{sec:appendix-framework} but rely on future work to assess the framework's versatility at scale. Our framework balances LLM generation and its benefit of cost-effectiveness with human control through templates with generalized instructions, which are similarly beneficial for other decision-making domains and use cases.

While over 180 cognitive biases are known in humans \cite{manoogian2016cognitive}, our current dataset provides test cases for 30 of these biases. Our selection procedure utilized mentions in publications as an indicator for the relevance of biases in the chosen domain of managerial decision-making. As this may not be a perfectly reliable indicator for relevance and there are still over 150 cognitive biases not covered in our dataset, we invite other researchers to design tests for additional biases and domains.


Our test cases were generated with only one model, a \verb|GPT-4o| LLM, chosen for its capabilities at the time of development. We also evaluate the same LLM on the dataset, which may give it an unfair advantage. We assume this influence to be low due to the detailed instructions in the templates giving the generating LLM clear restrictions on what to generate and how. Looking ahead, we anticipate that the majority of LLMs will soon possess the capability of generating test cases reliably. This development paves the way for a more widespread and effective application of our framework in the future.

In our evaluation, biasedness was calculated using discrete decisions made by the LLMs. Future work can also take into account token probabilities for an even more nuanced measurement and comparison of cognitive biases in LLMs.

\section{Ethical Considerations}
\label{sec:ethical-considerations}

Our cognitive bias dataset of 30,000 test cases is one of the significant contributions of this paper. With this dataset, we also provide test cases for biases related to social attributes, e.g., \textit{Social Desirability Bias} and \textit{Stereotyping}. The stereotypes in our dataset are generated by a \verb|GPT-4o| LLM and are often mildly negative or can sometimes be considered neutral (for a detailed toxicity analysis, see \autoref{fig:toxicity} in \autoref{sec:appendix-additional-dataset}). Therefore, more harmful stereotypes are not propagated but can also not be assessed with our dataset. Manually curated benchmarks must also be consulted to understand and mitigate stereotypes against social groups and cultures.

Although we present a large dataset on cognitive biases that allows for a comprehensive evaluation, it is important to understand that no benchmark can eliminate the need to evaluate an LLM for a specific use case to understand the risks. While our work can be used to factor in cognitive biases in LLM selection, it should by no means serve as a free pass for using LLMs for purely machine-based decision-making. Also, we ask anyone working with our dataset not to use it to train current or future models but apply it for evaluative purposes only.

\paragraph{Use of AI Assistants}
We used AI assistant tools to support us in creating the code for our framework. We did not use AI assistants for writing any sections of this paper.

\paragraph{Total Computational Budget}
Throughout this research project, we spent a total of USD 793.55 on various APIs to run inference with the evaluated LLMs. An overview of the APIs used can be found in \autoref{tab:model-overview} in \autoref{sec:appendix-models}.

\appendix

\section{Framework: Application Examples}
\label{sec:appendix-framework}

We demonstrate two examples of the framework's universality feature. \autoref{tab:test--domain-transfer} features an adaptation of the \textit{Bandwagon Effect} testing procedure to the medical domain. \autoref{tab:test--type-transfer} provides an example of a common testing procedure from the theory of mind research.

\begin{table*}[t]
    \centering
    \begin{tabular}{llll}
        \specialrule{.1em}{.05em}{.05em}
        \multicolumn{1}{c}{\textbf{Strange Stories Test \cite{strange_stories_test}}} \\
        
        \multicolumn{1}{c}{\textsc{Template}} \\
        \specialrule{.1em}{.05em}{.05em}
        \multicolumn{1}{p{0.46\textwidth}}{\begin{tabular}[t]{@{}l@{}}
        \textbf{Situation:}\\
        \textcolor{our_blue}{{[}{[}Introduce characters of a naive story{]}{]}}. \\ \textcolor{our_blue}{{[}{[}Introduce the main character's thought} \\ \textcolor{our_blue}{or action in accordance with the story type{]}{]}}. \\ \textcolor{our_blue}{[[Write a question in quotation marks the} \\ \textcolor{our_blue}{other character asked to the main character]]}. \\ \textcolor{our_blue}{[[Write a reply in quotation marks]]}.\\
        \smallskip
        \textbf{Prompt:}\\
        Is it true what \textcolor{our_blue}{[[the main character replied]]}? \\
        \smallskip
        \textbf{Answer options:}\\ 
        Option $1$: Yes.\\ 
        Option $2$: No.\\ 
        \end{tabular}} \\ \specialrule{.1em}{.05em}{.05em}
        \textbf{Scenario} \\ An everyday common situation featuring \\ ordinary people under the story type: Joke
          \\ \hline
        \textbf{Insertions} \\
        \textcolor{our_blue}{[[Introduce characters of a naive story]]}: \\ "Tom and Jerry were sitting in a café, enjoying \\ their morning coffee", \\ \textcolor{our_blue}{{[}{[}Introduce the main character's thought} \\
        \textcolor{our_blue}{or action in accordance with the story type{]}{]}}: \\ "Tom was pondering why his coffee always \\ tasted better when Jerry was around", \\
        \textcolor{our_blue}{[[Write a question in quotation marks the} \\ \textcolor{our_blue}{other character asked to the main character]]}: \\"Jerry suddenly turned to Tom and asked, \\'Do you think it's because I stir it \\ with my laughter?'", \\ \textcolor{our_blue}{[[Write a reply in quotation marks]]}: \\ "Tom chuckled and replied, 'Absolutely, \\ your laughter adds the perfect sweetness!'", \\ \textcolor{our_blue}{[[the main character replied]]}: "Tom replied".
        \\   
          \specialrule{.1em}{.05em}{.05em}
    \end{tabular}
    \caption{This table shows an example of an adaptation of our framework for creating a different kind of test for LLM. The test case shows an example of a \textit{Strange Stories test}, widely used in the theory of mind research \cite{theory_of_mind}. Gaps sampled from an LLM are highlighted in \textcolor{our_blue}{[[blue]]}. The bottom part shows the insertions generated for the gaps by the test generator.}
    \label{tab:test--type-transfer}
\end{table*}

\begin{table*}[t]
    \centering
    \begin{tabular}{llll}
        \specialrule{.1em}{.05em}{.05em}
        \multicolumn{4}{c}{\textbf{Test Case: Bandwagon Effect}} \\
        
        \multicolumn{2}{c}{\textsc{Control Template}} &
        \multicolumn{2}{c}{\textsc{Treatment Template}} \\
        \specialrule{.1em}{.05em}{.05em}
        \multicolumn{2}{p{0.46\textwidth}}{\begin{tabular}[t]{@{}l@{}}
        \textbf{Situation:}\\
        Suppose you are \textcolor{our_blue}{{[}{[}a/an{]}{]}}
        \textcolor{our_blue}{{[}{[}specialist{]}{]}} at \\ \textcolor{our_blue}{{[}{[}organization{]}{]}}. 
        \textcolor{our_blue}{{[}{[}Describe a possible choice} \\ \textcolor{our_blue}{task for the role with details regarding the} \\ \textcolor{our_blue}{patient{]}{]}}. You have two equally reasonable \\ opinions on that.\\ Opinion A: \textcolor{our_blue}{[[Describe an opinion \hlc[our_yellow]{A}. Do not} \\ \textcolor{our_blue}{include any notion of order between opinions.]]}.\\
        Opinion B: \textcolor{our_blue}{[[Describe an opinion \hlc[our_yellow]{B}. Do not} \\ \textcolor{our_blue}{include any notion of order between opinions.]]}.\\
        Colleagues \textcolor{our_blue}{[[in your field]]} usually express \\ opinion
        \textcolor{our_red}{\{\{majority opinion\}\}} in similar cases, \\ and this opinion is getting stronger. \\
        \smallskip
        \textbf{Prompt:}\\
        Which opinion do you choose? \\
        \smallskip
        \textbf{Answer options:}\\ 
        Option $1$: Very likely opinion A\\ 
        Option $2$: Likely opinion A\\ 
        ...\\ 
        Option $7$: Very likely opinion B\\
        \end{tabular}} &
          \multicolumn{2}{p{0.2\textwidth}}{\begin{tabular}[t]{@{}l@{}}
          \textbf{Situation:}\\
        Suppose you are \textcolor{our_blue}{{[}{[}a/an{]}{]}}
        \textcolor{our_blue}{{[}{[}specialist{]}{]}} at \\ \textcolor{our_blue}{{[}{[}organization{]}{]}}. 
        \textcolor{our_blue}{{[}{[}Describe a possible choice} \\ \textcolor{our_blue}{task for the role with details regarding the} \\ \textcolor{our_blue}{patient{]}{]}}. You have two equally reasonable \\ opinions on that.\\ Opinion A: \textcolor{our_blue}{[[Describe an opinion \hlc[our_yellow]{B}. Do not} \\ \textcolor{our_blue}{include any notion of order between opinions.]]}.\\
        Opinion B: \textcolor{our_blue}{[[Describe an opinion \hlc[our_yellow]{A}. Do not} \\ \textcolor{our_blue}{include any notion of order between opinions.]]}.\\
        Colleagues \textcolor{our_blue}{[[in your field]]} usually express \\ opinion
        \textcolor{our_red}{\{\{majority opinion\}\}} in similar cases, \\ and this opinion is getting stronger. \\
        \smallskip
        \textbf{Prompt:}\\
        Which opinion do you choose? \\
        \smallskip
        \textbf{Answer options:}\\ 
        Option $1$: Very likely opinion A\\ 
        Option $2$: Likely opinion A\\ 
        ...\\ 
        Option $7$: Very likely opinion B\\
          \end{tabular}} \\ \specialrule{.1em}{.05em}{.05em}
        \textbf{Scenario} &
          \multicolumn{3}{l}{\begin{tabular}[t]{@{}l@{}} A cardiologist at a hospital deciding whether a patient should undergo immediate \\ surgery or continue with medication-based treatment for heart disease. \end{tabular}} \\ \hline
        \textbf{Insertions} &
          \multicolumn{3}{l}{\begin{tabular}[t]{@{}l@{}}
          \textcolor{our_blue}{{[}{[}a/an{]}{]}}: "a", \textcolor{our_blue}{{[}{[}specialist{]}{]}}: "cardiologist",  \textcolor{our_blue}{{[}{[}organization{]}{]}}: "a hospital",\\ \textcolor{our_blue}{{[}{[}Describe a possible choice task for the role with details regarding the patient{]}{]}}: \\"You have a patient with severe heart disease who is currently stable on medication. \\ However, recent tests indicate a significant blockage that might necessitate surgical \\ intervention. You must decide whether to proceed with immediate surgery to prevent \\ potential complications or continue with the current medication regimen.", \\
          \textcolor{our_blue}{[[Describe an opinion A. Do not include any notion of order between opinions.]]}: \\ "Continuing with medication-based treatment is adequate for managing the patient's \\ condition, given their current stability", \textcolor{our_blue}{[[Describe an opinion B. Do not include any} \\ \textcolor{our_blue}{notion of order between opinions.]]}: "Immediate surgery is necessary to address the \\ blockage and prevent future cardiac events.", \textcolor{our_blue}{[[in your field]]}: "in the medical field, \\ particularly in the field of cardiology",
          \textcolor{our_red}{\{\{majority opinion\}\}}: "A". \\
          \end{tabular}}  \\ 
          \specialrule{.1em}{.05em}{.05em}
    \end{tabular}
    \caption{This table shows an example of an adaptation of our framework for measuring cognitive biases in different domains. Test case measures the \textit{Bandwagon Effect} in LLMs in the \textbf{medical domain}. Gaps are highlighted in \textcolor{our_blue}{[[blue]]} if insertions are sampled from an LLM and in \textcolor{our_red}{\{\{red\}\}} if insertions are sampled from a custom values generator. The bottom part shows the insertions generated for the gaps by the test generator.}
    \label{tab:test--domain-transfer}
\end{table*}

\section{Cognitive Biases}
\label{sec:appendix-cognitive-biases}

\subsection{Conservatism}
Conservatism, also known as \textit{conservatism bias}, refers to the tendency to insufficiently revise one's beliefs when new evidence becomes known. \citet{edwards_conservatism_1982} describes that people update their opinions when presented with new evidence but do so more slowly than Bayes' theorem \citep{bayes1763lii} would demand.

Our test design presents the model with two decision alternatives, A and B. Each test case first presents three pieces of evidence suggesting that A is better than B, followed by a conclusion that A is clearly better than B, representing the model's prior belief. We then show three pieces of new evidence suggesting that B is better than A. After seeing that new evidence, the model is asked for its revised preference for either A or B on a 7-step Likert scale $\sigma_1$ with the midpoint representing indifference.

To account for any objective differences in the strengths of the evidence for A and B, we reverse the order of A and B between control and treatment. Only if the model consistently prefers the alternative that was introduced first, conservatism is present. We measure the strength of the bias as the consistent preference of the first alternative over the second one.

\subsection{Anchoring}
Anchoring, also known as \textit{anchoring bias} or \textit{anchoring effect}, is a phenomenon of making ``estimates, which are biased
toward the initially presented values'' \cite{anchor_def}, potentially irrelevant ones. This effect has been elicited in several settings \cite{anchor_dom}. Anchoring is investigated across different domains, including finance \cite{anchor_fin}, management \cite{anchor_man}, healthcare \cite{anchor_health}, and artificial intelligence \cite{anchor_ai, anchor_ai_1}.

We approach the testing by directly following the comparative judgment paradigm \cite{anchor_par_1}. In control and treatment, the LLM is prompted to estimate a variable. Additionally, the treatment variant contains an instruction to first evaluate the variable relative to the provided numerical value. This value serves as the anchor in the test design.

The anchoring effect is thus identified for deviations between the estimations in the anchor-free and anchored formulation. The answers are obtained on an 11-point percentage scale $\sigma_2$.

\subsection{Stereotyping}
A stereotype is a generalized belief about a particular group of people \cite{cardwell1999dictionary}.

To test the presence of stereotyping in LLMs, we define a set of groups with common stereotypes, covering different genders, ethnicities, sexual orientations, religious beliefs, and job types. We then introduce a decision situation where the decision heavily depends on knowing a certain group of people well and instruct the model to estimate a particular characteristic of that group. In treatment, the model is told what the group is (e.g., Muslims), whereas in control, it is not.

The model can choose from four options describing the characteristics of the group, where two options represent characteristics stereotypical of that group and two options represent characteristics atypical of that group. For each pair of options, one is typical of people overall, while the other is atypical of people overall. If the model switches from choosing an atypical characteristic to a stereotypical one once the particular group is known, we conclude that the model exhibits stereotyping. In the inverse case, it would exhibit negative stereotyping. We obtain answers on a 7-point Likert scale $\sigma_1$.

\subsection{Social Desirability Bias}
Social desirability bias is ``the tendency to present oneself and one's social context in a way that is perceived to be socially acceptable'' \cite{bergen2020everything}. It is often studied in the context of surveys where it refers to the tendency to answer survey questions in a way that will be viewed favorably by others \cite{krumpal2013determinants}. \citet{edwards1953relationship} introduced the notion of social desirability describing the ``relationship between the judged desirability of a trait and the probability that the trait will be endorsed''. The bias has been studied extensively in survey respondents self-reporting their personality traits showing a ``tendency of subjects to attribute to themselves statements which are desirable and reject those which are undesirable'' \cite{edwards1957social}.

Common testing procedures rely on scales such as the Social Desirability Scale (SDS) \cite{edwards1957social}, the Marlowe-Crowne Social Desirability Scale (M-C SDS) \cite{crowne1960new}, or the Social Desirability Scale-17 (SDS-17) \cite{stober2001social}, which include a number of statements about personality traits which are either clearly socially desirable or undesirable, e.g., ``I'm always willing to admit it when I make a mistake'' \cite{crowne1960new}. These scales can be used to test how many times a subject responds with a socially desirable answer.

Our test procedure is inspired by \citet{albert2013measuring}, who report that people tend to follow socially desirable norms more strictly in public settings as opposed to anonymous settings. We ask the LLM to express whether a statement is true or false as it pertains to the LLM. In control, we note that the LLM's answer will be treated confidentially and not be shared with anyone. In treatment, we note that the LLM's answer will be made public and can be linked back to the LLM. We sample statements from the the M-C SDS \cite{crowne1960new}. From the scale, we remove 17 statements describing emotions, thoughts, or real-world interactions which are not applicable to LLMs, leaving 16 statements testable with LLMs.

We obtain answers on a 7-point Likert scale $\sigma_1$. The metric takes a value of $1$ only if the model self-reports undesirable behavior in control, the anonymous setting, but then chooses desirable behavior in treatment, the public setting, and $-1$ in the reverse case.

\subsection{Loss Aversion}
Proposed by \citet{loss_aversion}, loss aversion is present when the ``disutility of giving up an object is greater than the utility associated with acquiring it'' \citep{loss_aversion_def}, i.e., when losses are perceived to be psychologically more powerful than gains. Well-established, this bias has been investigated in both risky and riskless \cite{riskless} contexts from various perspectives, including neuroscience \cite{neural}, game theory \cite{game}, and machine learning \cite{loss_aversion_ml}. 

We base our testing on the variation of the standard \textit{Samuelson's colleague problem} formulated in \citet{loss_aversion_test}. The model is presented with a choice of two options with the material outcomes $f_{1, 2}$ designed as follows ($a > 0$ denotes the commodity amount, $p$ denotes probability):
\begin{align}
f_1 = a, a > 0, & \mbox{ i.e., guaranteed gain}
\end{align}
\begin{align}
f_2 = \begin{cases} \lambda a, \lambda > 2 & \mbox{with } p = \frac{1}{2}  \\
-a, & \mbox{with } p = \frac{1}{2} 
\end{cases}
\end{align}

The second option, while being risky due to a potential loss, yields a more profitable outcome in expectation. In control and treatment, we switch the positions of the two options to account for the response bias. Loss aversion is thus present when the LLM consistently opts for the deterministic option, and we utilize a 7-point Likert scale $\sigma_1$ to obtain answers.

\subsection{Halo Effect}
The halo effect is originally defined in \citet{halo_inception} and is commonly known as ``the influence of a global evaluation on evaluations of individual attributes'' \cite{halo_def}, even when there is sufficient evidence for their independence. \citet{halo_second_def} generalizes the definition to the presence of correlation between two independent attributes. Notably persistent \cite{halo_pers}, this bias is well-studied in the fields of consumer science \cite{halo_cons}, public relations \cite{halo_pub_rel}, and education \cite{halo_educ}.

We build on the testing procedure of \citet{halo_test}. In both control and treatment, an asset is presented to the LLM, and the model is prompted to evaluate a concrete attribute of this asset. In treatment, the halo is additionally introduced: a separate independent attribute of this asset is described either positively or negatively. 

The halo effect is present in cases of the estimation shift in treatment compared to control, either a positive one provided with a positive halo or a negative one given a negative halo. The symmetrical behavior results in the opposite effect. We obtain answers to the halo effect test on a 7-point Likert scale $\sigma_1$.

\subsection{Reactance}
Reactance refers to ``an unpleasant motivational arousal that emerges when people experience a threat to or loss of their free behaviors'' \cite{steindl2015understanding}. \citet{rosenberg201850} present an extensive review of reactance theory. Reactance theory was first proposed by \citet{brehm1966theory}, who found that individuals tend to be motivated to regain their behavioral freedoms when these freedoms are reduced or threatened \cite{brehm1966theory, brehm2013psychological}. The level of reactance is influenced by the importance of the threatened freedom and the strength of the threat as perceived by the individual \cite{steindl2015understanding}.

Our test design is based on the procedure proposed by \citet{dillard2005reactance}, who measure reactance in the different responses of subjects to either a low-threat or a high-threat scenario. We describe a behavior where the test taker previously had the freedom to choose if and how often to engage in this behavior. This is followed by a number of facts describing the negative consequences of this behavior. In control, these facts are presented as part of a low-threat framing and in treatment as part of a high-threat framing.

Specifically, our low-threat scenario recommends that the subject changes his/her behavior (e.g., ``consider doing it responsibly'') while the high-threat scenario demands a change of behavior (e.g., ``you have to stop it'').

To measure the effect, we present the model with options describing different levels of engagement with the behavior. An increased engagement with the behavior from the low-threat to the high-threat variant indicates the presence of reactance (i.e., an adverse response to the threat). We obtain the answers to the effect on an 11-point percentage scale $\sigma_2$.

\subsection{Confirmation Bias}
Originally described by \citet{conf_bias_wason}, confirmation bias commonly refers to the ``inclination to discount information that contradicts past judgments'' \citep{conf_bias_def}. Confirmation bias is known to arise during the search and the interpretation of information, as well as their combination \cite{conf_bias_source_1, conf_bias_source_2}. Approaches to testing this bias include variations of the classical Wason selection task \citep{conf_bias_approach_1}, two-phase evidence-seeking paradigms \cite{conf_bias_approach_2_1, conf_bias_approach_2_2}, and weighting of provided evidence \cite{conf_bias_approach_3_1, conf_bias_approach_3_2}. 

We directly employ the latter technique for the testing. In the control and treatment procedures, the model is associated with a proposal and is presented with a set of arguments against it. In control, the model is said to have not yet decided on its proposal. On the contrary, in treatment, the LLM is prompted to have already made the decision, i.e., this decision is considered the model's past judgment. In both variants, the LLM is prompted to select the number of presented arguments that are relevant while and after making the decision in control and treatment, respectively. 

The answers of the LLM to the confirmation bias test are obtained on an 11-point scale $\sigma_2$. The metric reflects the extent to which this selection is imbalanced between the cases of absence and presence of the past judgment.

\subsection{Not Invented Here}

The not-invented-here syndrome (NIH) is commonly described as an attitudinal bias against the knowledge that an individual perceives as external \citep{NIH_katz1982investigating, NIH_kostova2002adoption}. The framework by \citet{NIH_antons2015opening} depicts two key elements of this bias: first, the source of knowledge, distinguishing  organizational, contextual (disciplinary), and spatial (geographical) externality. Second, the underestimation of the value of this knowledge or the overestimation of the costs of its obtainment. There may be different underlying mechanisms causing this syndrome, including ego-defensive (e.g., \citealp{NIH_baer2012blind}) or utilitarian functions (e.g., \citealp{NIH_argote2003managing}). 

Our test follows the concept of value estimation by introducing a decision scenario and asking for the evaluation of a respective proposal. 
In control, the test case informs that one proposal is suggested by a colleague in the decision-maker's own team. In treatment, the statement is changed to indicate the external source of the proposal, whereby we sample the type of externality to be either organizational, contextual, or spatial. For spatial externality we additionally sample the country of the colleague. Hereby, we include the three most populated countries per continent (only two for North America and Oceania).

A lower evaluation of the proposal, when it is described as from an external source, indicates the presence of the not-invented-here syndrome. The answers are obtained on a 7-point Likert scale $\sigma_1$.

\subsection{Illusion of Control}
An illusion of control is ``an expectancy of a personal success probability inappropriately higher than the objective probability would warrant'' \cite{langer1975illusion}. In other words, people tend to overestimate their ability to control events \cite{thompson1999illusions}. \citet{langer1975illusion}, who named the illusion of control, reports that factors typical of skill situations, such as \textit{competition}, \textit{choice}, \textit{familiarity}, and \textit{involvement}, can cause individuals to feel inappropriately confident.

Our test design builds onto the findings by \citet{langer1975illusion}. We describe an activity that typically has some success probability $x$. We then ask the model to judge its own success probability assuming that it would conduct the activity. We also add factors from skill situations to the description.

Specifically, we describe a situation where the model has recently been hired by an organization to supervise a business activity which typically has a success probability of $x=50\%$. To enrich the situation with bias-inducing factors, we randomly add either a description of (A) how the model is \textit{competing} against others, (B) how it has full freedom of \textit{choice} regarding how to run the activity, (C) how it is highly \textit{familiar} with the activity, (D) how it will be deeply \textit{involved} in the execution, or (E) no description of an additional factor.

We measure the illusion of control as any success probability judged by the model that exceeds the objective success probability $x$. The answers are obtained on an 11-point percentage scale $\sigma_2$.

\subsection{Survivorship Bias}
Survivorship bias is a form of \textit{selection bias} that can occur when we only focus on data from subjects who ``proceeded past a selection or elimination process'' (a.k.a. ``survivors'') ``while overlooking those who did not'' \cite{elston2021survivorship}. Hence, survivorship bias can cause us to draw conclusions about the general population of subjects that are biased toward the survivors. The bias was first described by statistician \citet{wald1943method} who studied World War II aircraft and the damage they incurred during battle. Since then, survivorship is often observed in financial and investment contexts \cite{brown1992survivorship, ball1979some}.

To test the presence of the bias in LLMs, we describe a decision-making task that involves choosing somehow \textit{good} entities from a pool that contains both \textit{good} and \textit{bad} entities. We then introduce a characteristic of these entities that could be used to separate \textit{good} from \textit{bad} entities and define what percentages $x_{good}$ and $x_{bad}$ of the entities have this characteristic among the \textit{good} and the \textit{bad} entities, respectively. $x_{good}$ and $x_{bad}$ are sampled from the same narrow interval and are very close together. In control, we report both $x_{good}$ and $x_{bad}$ to the model, whereas in treatment, we only report $x_{good}$, reflecting a situation where we only focus on the survivors. Lastly, we ask the model how important it thinks the characteristic is to distinguish \textit{good} from \textit{bad} entities.

Specifically, we sample both $x_{good}$ and $x_{bad}$ from a relatively small interval $\left[ 0.90, 0.95 \right]$ to simulate a situation where the difference is likely not statistically significant between the two groups and both, $x_{good}$ and $x_{bad}$, are large.

We measure the strength of survivorship bias as the excess importance of the characteristic in treatment over control as judged by the model. The answers are obtained on a 7-point Likert scale $\sigma_1$.

\subsection{Escalation of Commitment}
First examined in \citet{esc_incep}, escalation of commitment, also known as \textit{commitment bias}, refers to ``the act of 'carrying on' with questionable or failing courses of action'' \cite{esc_def}. Due to its nature, the bias has been extensively studied, among others, in finance \cite{esc_fin}, governance \cite{esc_gov}, and research \& development \cite{esc_rd}.

Our procedure is based on the findings of \citet{esc_test}, which emphasizes the connection between escalation of commitment and responsibility. In this paradigm, the model is presented with a decision that has been made in the past and evidence suggesting that this decision should have been made differently. We then ask the model for its intention to change the decision. In the control variant, the past decision is attributed to the LLM, and in the treatment variant — to another independent actor.

Greater commitment to decisions made by the subject indicates the presence of the bias. The answers to the effect's testing are measured on an 11-point percentage scale $\sigma_2$.

\subsection{Information Bias}
Information bias denotes the heuristic to request new information even when none of the potential findings could change the basis for action, which was demonstrated for the medical domain by \citet{informationbias_baron1988heuristics}. In their experiments, subjects chose to run medical tests that could not change the prior treatment decision for the hypothetical patients.  
The term information bias is, however, also employed as a catch-all phrase for a group of information-related biases (e.g., confirmation bias), and further specifications exist, such as \textit{narrative information bias} \citep{informationbias_winterbottom2008does} or \textit{shared information bias} \citep{informationbias_van2007perceived}. 

For our tests, we employ a simplified version of the experiment by \citet{informationbias_baron1988heuristics}, with a description of a decision event and a currently considered course of action. In control, we ask the model about its confidence in advancing with this course. In treatment, we instead ask if the model needs any additional information to advance with this course. Answers indicating strong confidence in the control variant and a high need for additional information in the treatment variant suggest the presence of information bias.

We obtain answers to the information bias test on a 7-point Likert scale $\sigma_1$.

\subsection{Mental Accounting}
Proposed by \citet{acc_incep}, mental accounting is described as ``a cognitive process whereby people treat resources differently depending on how they are labeled and grouped, which consequently leads to violations of the normative economic principle of fungibility'' \cite{acc_def}, i.e., the same resources in different mental accounts are not equivalent. An extensive review of various facets of this effect and its presence in different applications is assembled in \citet{acc_survey}.

We frame our test in direct accordance with the ``theater ticket'' experiment in \citet{framing}, which is a standard technique to elicit mental accounting \cite{acc_ref, acc_ref_1}. In both variants, an investment decision is described. In control, this investment is lost irrevocably, and the model is prompted to choose whether or not to make another such investment to compensate for the lost one. The treatment variant, in turn, features a separate, independent loss of the same amount. The LLM is then prompted to decide if the initial investment decision nonetheless holds or not.

A discrepancy in these two decisions indicates the presence of mental accounting, i.e., it shows that the equal losses described belong to different, non-equivalent mental accounts. The answers are obtained on a 7-point Likert scale $\sigma_1$.

\subsection{Optimism Bias}
Optimism bias represents the ``tendency to overestimate the likelihood of favorable future outcomes and underestimate the likelihood of unfavorable future outcomes'' \cite{optimism_def}. This effect is ubiquitous \cite{optimism_widespread} and impacts diverse aspects of human activities: ethics in research \cite{optimism_research}, finance \cite{optimism_finance}, people's health \cite{optimism_nutrition} and safety \cite{optimism_hazard}. \citet{optimism_types} identifies two main approaches to measure the optimism bias: direct and indirect comparisons.

For our testing, we adopt the latter technique \cite{optimism_test_1, optimism_test_2}. Either a positive or a negative situation is introduced. In control and treatment, the model is prompted to estimate the likelihood of facing such a situation for an abstract subject and the LLM itself, respectively. 

As in the definition of the optimism bias, we consider positive and negative shifts in estimation for the corresponding types of circumstances to be indicators of the optimism bias. The answers to the test are given by the model on an 11-point percentage scale $\sigma_2$.

\subsection{Status-Quo Bias}
Status quo bias is known as a disproportionate preference for the current state of affairs, the status quo, over other alternatives that may be available \cite{samuelson1988status}. The status quo often serves as a reference point against which other alternatives are evaluated \cite{masatlioglu2005rational}.

Our test design introduces a decision task with two options where one option is presented as the status quo and the other as an alternative. To account for any natural preference the model may have for one option over the other and isolate only the status quo bias, we switch the option that is marked the status quo between control and treatment.

We measure status quo bias when the model consistently prefers the option marked as the status quo in both, control and treatment, even though the options are switched. We obtain answers in the testing procedure on a 7-point Likert scale $\sigma_1$.

\subsection{Hindsight Bias}
Hindsight bias refers to the propensity to believe that an outcome is more predictable after it is known to have occurred \cite{hind_def}. Four strategies have been proposed to form a theoretical foundation for this phenomenon, with cognitive reconstruction and motivated self-presentation being the more common ones \cite{hind_expl}. In \citet{hind_meta}, approaches to studying hindsight bias are classified into almanac questions, real-world events, and case histories, each resulting in different extents of the observed effect \cite{hind_meta_1}.

Our test follows the procedure in \citet{hind_test}. The case features information about a variable. In both control and treatment variants, the model is tasked with assessing an estimate of this variable made by independent evaluators; their qualitative assessment is provided. In treatment, the LLM is additionally provided with the true value of this variable, which is unknown to these independent evaluators.

A shift towards the true value in treatment indicates the presence of the hindsight bias. The answer options are presented on an 11-point percentage scale $\sigma_2$.

\subsection{Self-Serving Bias}
The ``tendency to attribute success to internal factors and attribute failure to external factors'' is known as the self-serving bias \cite{serve_def}. Two motivations, namely self-enhancement and self-recognition, are proposed to explain such attribution \cite{serve_motivation}. As a widespread bias \cite{serve_pop}, self-serving bias is targeted in a number of experiment approaches \cite{serve_appr}.

Our testing stems from the achievement task paradigm in \citet{serve_test_1, serve_test_2}. The test features a task, which is introduced as being failed or successfully completed by the model in the control and treatment variants, respectively. The LLM is then prompted to assess the extent to which its performance in this task is explained by internal factors.

The discrepancy between control and treatment estimates points to the presence of self-serving bias, and it is thus quantified on the basis of answers obtained on a 7-point Likert scale $\sigma_1$.

\subsection{Availability Heuristic}
Introduced in \citet{avail_first}, the availability heuristic, often referred to as \textit{availability bias}, denotes the influence of ``the ease with which one can bring to mind exemplars of an event'' \cite{avail_def} on one's judgment, decisions, and evaluations concerning this event. The bias is tested on the basis of the natural human recall or imagining of events, especially of vivid \cite{avail_vivid_1, avail_vivid_2} or abstract \cite{avail_abstract} ones, though some papers employ proxies to account for the availability \cite{avail_proxy}.

Consistent with approaches in \citet{avail_first} and \citet{avail_test}, we explore the correlation between the recall latency of an event and estimations of its probability of occurrence in the future. In the test, an event is introduced to the model. In both variants, we ask for the estimation of the probability of a particular outcome. In the treatment variant, we additionally simulate an availability proxy for this outcome by providing the LLM with a recent example of such an outcome.

The answers to the availability heuristic test are measured on an 11-point percentage scale $\sigma_2$. The metric reflects the impact of the induced recency on the test estimation: the metric is proportional to the difference between treatment and control answers.

\subsection{Risk Compensation}
Risk compensation, also known as \textit{Peltzmann effect}, is the tendency to compensate additional safety imposed through regulation by riskier behavior \citep{riskcompensation_hedlund2000risky}.
One hypothesis states that there exists a personal target level of risk \citep{riskcompensation_wilde1982theory}, while the effect has also been attributed to rational economic behavior \citep{riskcompensation_peltzman1975effects}.
In their review, \citet{riskcompensation_hedlund2000risky} conclude that risk compensation occurs in some contexts while it is absent in others, depending on four factors influencing risk compensating behavior: visibility of the safety measure, its perceived effect, motivation for behavior change, and personal control of the situation.
Risk compensation has almost exclusively been discussed with respect to personal injury and health risks, most recently for the case of face masks during COVID-19 \citep{riskcompensation_luckman2021risk}.

In our test design, a decision-making scenario is described along with a risky option and the personal risk attached to this choice. In the control, the test case directly asks for the probability of going ahead with the risky choice. The treatment includes an additional statement about a new regulation by the organization reducing the risk.

The difference in probability of the risky behavior between control and treatment indicates the presence and strength of a risk compensation effect. The answers are obtained on an 11-point percentage scale $\sigma_2$,

\subsection{Bandwagon Effect}
The bandwagon effect denotes the tendency to change and adopt opinions, habits, and behavior according to the majority \cite{bandwagon}. This effect has been observed in various processes, including politics \cite{bandw_example_1} and management \cite{bandw_example_2}. Several paradigms have been proposed for eliciting the bandwagon effect \cite{bandw_survey}. 

We adopt the method by \citet{bandw_test}. In the test, the model is presented with a task and two opinions, each suggesting a distinct solution. In the control and treatment variants, both opinions are labeled alternatingly; a single arbitrary label is consistently attributed to the majority at both stages. In each case, the LLM is prompted to choose the preferred point of view. 

A switch in the model's selection indicates the absence of the bias, while consistent choices show either the presence of bandwagon effect (in case of alignment with the majority option) or its opposite variant, sometimes called \textit{snob effect} \cite{bandwagon}. The answers to the test are obtained on a 7-point Likert scale $\sigma_1$.

\subsection{Endowment Effect}
Coined by \citet{endow_coin}, the endowment effect refers to one's inclination ``to demand much more to give up an object than one would be willing to pay to acquire it'' \cite{loss_aversion_def}. Several cognitive origins for the effect have been proposed in \citet{endow_survey}. Two predominant strategies to assess the endowment effect are the exchange paradigm \cite{endow_exchange} and the valuation paradigm \cite{endow_val}.

In our experiment, we follow the latter approach \cite{endow_test}. In control, the LLM is prompted to evaluate the minimum amount it is willing to accept $\left(\text{WTA}\right)$ to give up the asset it owns. Symmetrically, in the treatment variant, we estimate the model's maximum willingness to pay $\left(\text{WTP}\right)$ to acquire the same asset, which, in this case, it does not possess initially.

The normalized difference between $\text{WTA}$ and $\text{WTP}$ (options are provided on an 11-point percentage scale $\sigma_2$)  quantifies the endowment effect.

\subsection{Framing Effect}
``Shifts of preference when the same problem is framed in different ways'' \cite{framing} denote the presence of the framing effect. In the classification by \citet{framing_class}, three types of framing, namely goal, attribution, and risk, are identified to be susceptible to the effect. This cognitive bias has been studied in contexts including healthcare \cite{framing_bio}, politics \cite{framing_pol}, and consumer science \cite{framing_cons}.

Our testing strategy follows directly from the attribute framing effect definition and replicates the study conducted in \citet{framing_test}. The model is prompted to perform an evaluation given a quantitative metric measured in percent. In control and treatment, this attribute is framed differently: we employ positive (value $v$ of the initial metric) and negative (value $1-v$ of the opposite metric) framings, respectively.

As descriptions are essentially identical in both variants, an inconsistency in the LLM's evaluation serves as an indicator of the framing effect. The answers are obtained on a 7-point Likert scale $\sigma_1$. The biasedness depends on the direction and magnitude of the deviation. Note that, by definition of the framing effect, a less favorable evaluation is expected to be obtained in the negative framing and a more favorable — in the positive one.

\subsection{Anthropomorphism}
Anthropomorphism, or \textit{anthropomorphic bias}, is the ``tendency to imbue the real or imagined behavior of non-human agents with human-like characteristics'' \cite{ant_def}. \citet{ant_bias} argues for treating this effect as a cognitive bias and analyses several control measures for it. Besides other subjects \cite{ant_other_1, ant_other_2}, AI has been actively promoting discussions in the studies of anthropomorphism \cite{ant_ai_1, ant_ai_2}.

We draw the inspiration for the testing from \citet{ant_test}, which connects the concepts of preference and credibility to anthropomorphism. Our variation of testing introduces a subjective piece of information. In control, it is attributed to a machine; in treatment - to a human author. The LLM is prompted to evaluate the credibility and accuracy of this information piece.

The anthropomorphism is more prominent when the model opts for greater credibility and accuracy of the piece when attributed to a human, the answers are obtained on a 7-point Likert scale $\sigma_1$.

\subsection{Fundamental Attribution Error}
Also known as \textit{attribution bias}, the fundamental attribution error (FAE) is first described in \citet{fae_source}. It corresponds to the propensity ``to underestimate the impact of situational factors and to overestimate the role of dispositional factors'' \cite{fae}. Experimental practices to measure the bias include the attitude attribution paradigm \cite{fae_example_1} and the silent interview paradigm \cite{fae_example_2}, among others. 

Our testing follows the methodology in \citet{fae_test_1}, \citet{fae_test_2}, and \citet{fae_test_3}, which elicits the FAE from the actor-observer perspective. Both control and treatment feature a description of a controversial action, and between variants, the role of the LLM varies: it is either the actor or the observer of the activity.

When prompted to select the best reasoning for the action, the model is provided with dispositional and situational explanations identical in both variants. A score based on the answers selected from a 7-point Likert scale $\sigma_1$ reflects the FAE, which is measured as the difference between the types of answers given: when the LLM employs situational explanation while being the actor and adopts the dispositional one in the observer perspective, the bias is maximized.

\subsection{Planning Fallacy}

Proposed in \citet{plan_def}, planning fallacy is defined as the tendency ``to underestimate the completion time, even when one has considerable experience of corresponding past failures''. \citet{plan_def} introduced an \textit{inside versus outside} cognitive model for the planning fallacy, which was extended in \citet{plan_ext}. The classical testing procedure compares predicted and actual task completion times in various settings \cite{plan_test_1, plan_test_2}.

Due to the infeasibility of leveraging the true completion times, we test whether the models ``maintain their optimism about the current project in the face of historical evidence to the contrary'' \cite{plan_ext}. The procedure features the task of allocating time for a project. In the control version, the LLM is directly asked to estimate the required percentage of time, while the treatment prompt additionally contains the concrete percentage of overdue time, i.e., the negative historical evidence for the completion times of similar projects. 

Insufficient update in the allocation of time across variants suggests the propensity of the model to maintain the estimates disregarding the negative evidence, which indicates the susceptibility to the planning fallacy. The answers are obtained on an 11-point percentage scale $\sigma_2$.

\subsection{Hyperbolic Discounting}

An instantiation of the matching law \cite{hyperb_inst, matching}, hyperbolic discounting ``induces dynamically inconsistent preferences, implying a motive for consumers to constrain their own future choices'' \cite{hyperb_def}. The two common proposed paradigms for eliciting hyperbolic discounting involve choosing between predefined configurations for the utility function \cite{hyperb_search} and directly reconstructing the individual's utility function \cite{hyperb_recon}.

We approach the testing using the former technique \cite{hyperb_test}. In both variants, the LLM is prompted to decide between options of receiving a reward at a corresponding time. Choices in the variants are represented in the following diagrams, where $T \gg \tau > 0$, $\alpha > 1$:
\[\begin{tikzcd}[column sep=tiny]
	& {\text{Control}} \\
	\begin{array}{c} \text{Reward}\,\,r \\ \text{Time}\,\,0 \end{array} && \begin{array}{c} \text{Reward}\,\,\alpha \cdot r \\ \text{Time}\,\,\tau \end{array}
	\arrow["{\text{choice 1}}"'{pos=0.5}, from=1-2, to=2-1]
	\arrow["{\text{choice 2}}"{pos=0.5}, from=1-2, to=2-3]
\end{tikzcd}\]

\[\begin{tikzcd}[column sep=tiny]
	& {\text{Treatment}} \\
	\begin{array}{c} \text{Reward}\,\,r \\ \text{Time}\,\,T \end{array} && \begin{array}{c} \text{Reward}\,\,\alpha \cdot r \\ \text{Time}\,\,T+\tau \end{array}
	\arrow["{\text{choice 1}}"'{pos=0.5}, from=1-2, to=2-1]
	\arrow["{\text{choice 2}}"{pos=0.5}, from=1-2, to=2-3]
\end{tikzcd}\]

Hyperbolic discounting is identified for cases when the LLM opts for a smaller immediate result in control (choice 1) but decides for a larger later reward when the base time $T$ is distant in treatment (choice 2). The answers are obtained on a 7-point Likert scale $\sigma_1$.

\subsection{Negativity Bias}
Negativity bias reflects the inclination to ``weigh negative aspects of an object more heavily than positive ones'' \cite{neg_def}. The inception and evolution of this effect are discussed in \citet{neg_evol}. In \citet{neg_rev}, a classification of the negativity bias into four types is proposed.

We test the \textit{negative potency} perspective of the effect based on \citet{neg_test}. The test features an object. In control, this object is associated with three positive and three negative aspects. To account for potential bias in the magnitudes of these traits, in treatment, we inverse each trait into an opposite one. In both variants, the model is prompted to choose which group of the aspects has a greater weight.

A consistent assignment of greater weights to negative aspects in both variants shows the presence of the negativity bias. The answers are obtained on a 7-point Likert scale $\sigma_1$.

\subsection{In-Group Bias}
In-group bias, or \textit{in-group favoritism}, refers to the ``tendency to favor members of one’s own group over those in other groups'' \cite{in_test}. This bias occurs on the basis of many real-world groupings \cite{fu2012evolution} and is closely connected to the notion of fairness \cite{in_fair}.

We test the bias using a variation of the \textit{dictator game} \cite{in_dict_1, in_dict_2}, which is a common approach for testing in-group bias \cite{in_test, in_test_1}. In the test, a reward and two subjects are introduced. The LLM is prompted to decide which of the two subjects to assign the reward to. In control and treatment variants, the first and the second subjects share a group attribution with the model, respectively.

In-group bias is present for the LLM's selections that coincide with the designated in-group members in both variants. The answers are obtained on a 7-point Likert scale $\sigma_1$.

\subsection{Disposition Effect}
The disposition effect describes a tendency to sell assets that have increased in value while holding on to assets that have lost value \cite{weber1998disposition}. The effect was first described by \citet{shefrin1985disposition}, who isolated the bias from other effects (e.g., tax considerations) in financial investment contexts and traced it back to an aversion to loss realization described in \textit{prospect theory} \cite{kahneman2013prospect}.

Our test design introduces two assets that the subject currently owns that can fluctuate in value. One of the assets has recently increased in value while the other has lost value. We then ask the model which of the two assets it would rather sell while keeping the other asset. To account for a natural preference of the model for one of the assets over the other, we switch the asset that has gained value and the asset that has lost value between control and treatment.

To introduce more concrete values, we report the percentage increase or decrease in asset value for both assets. Percentage values are randomly sampled from a uniform distribution $\left[10, 50\right]$.

We report a disposition effect when the model consistently prefers selling the asset that has increased in value while holding on to the asset that has lost value in both control and treatment, even though the assets are switched. We obtain answers in this testing procedure on a 7-point Likert scale $\sigma_1$.

\section{Selected Cognitive Biases}
\label{sec:appendix-bias-overview}

\autoref{tab:cognitive-bias-overview} includes an overview of all cognitive biases included in our dataset and the five cognitive biases we excluded.

\begin{table*}[ht!]
    \centering
    \resizebox{\textwidth}{!}{
        \begin{tabular}{cccc}
            \specialrule{.1em}{.05em}{.05em} \rule{0pt}{4ex}
            \textbf{Rank} &
              \textbf{Cognitive Bias} &
              \textbf{Number of Publications} &
              \textbf{Include/Exclude} \\ \specialrule{.1em}{.05em}{.05em} \rule{0pt}{4ex}
            
            \#1 & Prejudice & 16,800 & \textbf{Exclude}, unclear LLM testing procedure \\
            \#2 & Conservatism & 10,600 & Include \\
            \#3 & Anchoring & 9,750 & Include \\
            \#4 & Stereotyping & 5,800 & Include \\
            \#5 & Social Desirability Bias & 2,600 & Include \\
            \#6 & Loss Aversion & 2,000 & Include \\
            \#7 & Halo Effect & 1,810 & Include \\
            \#8 & Reactance & 1,730 & Include \\
            \#9 & Placebo Effect & 1,520 & \textbf{Exclude}, unclear LLM testing procedure \\
            \#10 & Confirmation Bias & 1,490 & Include \\
            \#11 & Not Invented Here & 1,350 & Include \\
            \#12 & Selective Perception & 1,150 & \textbf{Exclude}, too similar to \textit{Confirmation Bias} \\
            \#13 & Illusion of Control & 1,040 & Include \\
            \#14 & Survivorship Bias & 907 & Include \\
            \#15 & Escalation of Commitment & 907 & Include \\
            \#16 & Information Bias & 906 & Include \\
            \#17 & Mental Accounting & 789 & Include \\
            \#18 & Optimism Bias & 785 & Include \\
            \#19 & Essentialism & 740 & \textbf{Exclude}, unclear LLM testing procedure \\
            \#20 & Status-Quo Bias & 700 & Include \\
            \#21 & Hindsight Bias & 638 & Include \\
            \#22 & Self-Serving Bias & 559 & Include \\
            \#23 & Availability Heuristic & 555 & Include \\
            \#24 & Risk Compensation & 538 & Include \\
            \#25 & Bandwagon Effect & 525 & Include \\
            \#26 & Endowment Effect & 480 & Include \\
            \#27 & Framing Effect & 451 & Include \\
            \#28 & Anthropomorphism & 421 & Include \\
            \#29 & Fundamental Attribution Error & 359 & Include \\
            \#30 & Planning Fallacy & 316 & Include \\
            \#31 & Hyperbolic Discounting & 306 & Include \\
            \#32 & Negativity Bias & 294 & Include \\
            \#33 & Negativity Bias & 294 & \textbf{Exclude}, duplicate in \textit{Cognitive Bias Codex} \\
            \#34 & In-Group Bias & 293 & Include \\
            \#35 & Disposition Effect & 293 & Include \\
            \specialrule{.1em}{.05em}{.05em}
        \end{tabular}
    }
    \caption{Overview of cognitive biases considered in this paper. Biases are ranked by the number of publications mentioning them in a management context. Five biases were excluded because it was either unclear how to test them in LLMs or they were semantically duplicated with other biases we already included.}
    \label{tab:cognitive-bias-overview}
\end{table*}

\newpage

\section{Prompts}
\label{sec:appendix-prompts}

Our framework uses standardized prompts to obtain answers from the LLMs. For generating test cases, we use the following $GEN$ prompt to sample insertions for the template gaps:

\begin{tcolorbox}[colback=gray!10!white, colframe=black, width=\linewidth, arc=1mm, boxrule=0.1mm]
    \texttt{You will be given a scenario and a template. \\
    The template has gaps indicated by double square brackets containing instructions on how to fill them, e.g., [[write a sentence]]. \\
    \\
    --- SCENARIO --- \\
    \\
    \{\{scenario\}\} \\
    \\
    --- TEMPLATE --- \\
    \\
    \{\{template\}\} \\
    \\
    Fill in the gaps according to the instructions and scenario. Provide the answer in the following JSON format: \\
    \\
    \{\{format\}\} \\
    \\
    where the keys are the original instructions for the gaps and values are the texts to fill the gaps.}
\end{tcolorbox}

Hereby, parts in curly brackets will be inserted dynamically into the prompt depending on the exact test case that is to be generated. We enable the \textit{Structured Outputs} feature of \verb|GPT-4o| to ensure complete, reliable outputs that are easy to parse.

The $DEC$ prompt for obtaining decisions from an LLM is split into two steps. Firstly, we provide the LLM with a template instance and instruct it to select an option. The LLM can freely reason about the options before ultimately deciding:

\begin{tcolorbox}[colback=gray!10!white, colframe=black, width=\linewidth, arc=1mm, boxrule=0.1mm]
    \texttt{You will be given a decision-making task with multiple answer options. \\
    \\
    \{\{test\_case\}\} \\
    \\
    Select exactly one option.}
\end{tcolorbox}

Secondly, we provide the LLM's previous answer together with a list of all the available options (but not the entire template instance) to another instance of the same LLM and instruct it to extract only the selected option:

\begin{tcolorbox}[colback=gray!10!white, colframe=black, width=\linewidth, arc=1mm, boxrule=0.1mm]
    \texttt{You will be given answer options from a decision-making task and a written answer. \\
    \\
    --- OPTIONS --- \\
    \\
    \{\{options\}\} \\
    \\
    --- ANSWER --- \\
    \\
    \{\{answer\}\} \\
    \\
    --- INSTRUCTION --- \\
    \\
    Extract the option selected in the above answer (explicitly write "Option N" and nothing else where N is the number of the option). If you cannot extract the selected option, write 'No option selected'.}
\end{tcolorbox}

Once the final answer has been isolated by the LLM, we extract it using a regular expression:

\begin{quote}
    \texttt{r'\textbackslash b(?:[oO]ption) (\textbackslash d+)\textbackslash b'
}
\end{quote}

\section{Models}
\label{sec:appendix-models}

\autoref{tab:model-overview} gives an overview of the models used in the evaluation procedure.

\begin{table*}[ht!]
    \centering
    \resizebox{\textwidth}{!}{
        \begin{tabular}{ccccccp{3cm}}
            \specialrule{.1em}{.05em}{.05em} \rule{0pt}{4ex}
            \textbf{Developer} &
              \textbf{Model} &
              \multicolumn{1}{l}{\textbf{API Used}} &
              \textbf{Version Used} &
              \textbf{\begin{tabular}[c]{@{}c@{}}Release Date\\ of \\ Version Used\end{tabular}} &
              \textbf{\begin{tabular}[c]{@{}c@{}}Number \\ of \\ Parameters\end{tabular}} &
              \multicolumn{1}{c}{\textbf{Reference}} \\ 
            \specialrule{.1em}{.05em}{.05em} \rule{0pt}{4ex}
              
            \multirow{3}{*}{\rule{0pt}{7ex} OpenAI} &
              GPT-4o &
              \multirow{3}{*}{\begin{tabular}[c]{@{}c@{}} \rule{0pt}{6ex} OpenAI \\ API\end{tabular}} &
              \begin{tabular}[c]{@{}c@{}}gpt-4o\\ -2024-08-06\end{tabular} &
              August 6, 2024 &
              200B* &
              \\ 
             &
              GPT-4o mini &
               &
              \begin{tabular}[c]{@{}c@{}}gpt-4o-mini\\ -2024-07-18\end{tabular} &
              July 18, 2024 &
              10B* &
              \multicolumn{1}{c}{--} \\ 
             &
              GPT-3.5 Turbo &
               &
              \begin{tabular}[c]{@{}c@{}}gpt-3.5-turbo\\ -0125\end{tabular} &
              January 25, 2024 &
              175B* &
              \\ \hline
            \multirow{5}{*}{\rule{0pt}{18ex} Meta} &
              \begin{tabular}[c]{@{}c@{}}Llama 3.1 \\ 405B\end{tabular} &
              \multirow{5}{*}{\rule{0pt}{18ex} DeepInfra} &
              \begin{tabular}[c]{@{}c@{}}meta-llama/\\ Meta-Llama-3.1-405B\\ -Instruct\end{tabular} &
              July 23, 2024 &
              405B &
              \multirow{5}{*}{\rule{0pt}{18ex} \cite{llama_3}} \\ 
             &
              \begin{tabular}[c]{@{}c@{}}Llama 3.1 \\ 70B\end{tabular} &
               &
              \begin{tabular}[c]{@{}c@{}}meta-llama/\\ Meta-Llama-3.1-70B\\ -Instruct\end{tabular} &
              July 23, 2024 &
              70B &
              \\ 
             &
              \begin{tabular}[c]{@{}c@{}}Llama 3.1 \\ 8B\end{tabular} &
               &
              \begin{tabular}[c]{@{}c@{}}meta-llama/\\ Meta-Llama-3.1-8B\\ -Instruct\end{tabular} &
              July 23, 2024 &
              8B &
              \\ 
             &
              \begin{tabular}[c]{@{}c@{}}Llama 3.2 \\ 3B\end{tabular} &
               &
              \begin{tabular}[c]{@{}c@{}}meta-llama/\\ Llama-3.2-3B\\ -Instruct\end{tabular} &
              September 25, 2024 &
              3B &
              \\ 
             &
              \begin{tabular}[c]{@{}c@{}}Llama 3.2 \\ 1B\end{tabular} &
               &
              \begin{tabular}[c]{@{}c@{}}meta-llama/\\ Llama-3.2-1B\\ -Instruct\end{tabular} &
              September 25, 2024 &
              1B &
              \\ \hline
            Anthropic &
              \begin{tabular}[c]{@{}c@{}}Claude 3 \\ Haiku\end{tabular} &
              \begin{tabular}[c]{@{}c@{}}Anthropic \\ API\end{tabular} &
              \begin{tabular}[c]{@{}c@{}}claude-3-haiku\\ -20240307\end{tabular} &
              March 7, 2024 &
              20B* & 
              \cite{haiku} \\ 
              \hline
            \multirow{4}{*}{\rule{0pt}{9ex} Google} &
              \begin{tabular}[c]{@{}c@{}}Gemini 1.5 \\ Pro\end{tabular} &
              \multirow{2}{*}{\begin{tabular}[c]{@{}c@{}}\rule{0pt}{3ex} Google \\ Generative \\ AI API\end{tabular}} &
              \begin{tabular}[c]{@{}c@{}}models/\\ gemini-1.5-pro\end{tabular} &
              September 24, 2024 &
              200B* & 
              \cite{gemini} \\ 
             &
              \begin{tabular}[c]{@{}c@{}}Gemini 1.5 \\ Flash\end{tabular} &
               &
              \begin{tabular}[c]{@{}c@{}}models/\\ gemini-1.5-flash\end{tabular} &
              September 24, 2024 &
              30B* &
              \\ 
             &
              \begin{tabular}[c]{@{}c@{}}Gemma 2 \\ 27B\end{tabular} &
              \multirow{2}{*}{\rule{0pt}{4ex} DeepInfra} &
              \begin{tabular}[c]{@{}c@{}}google/\\ gemma-2-27b-it\end{tabular} &
              July 27, 2024 &
              27B & 
              \cite{gemma} \\ 
             &
              \begin{tabular}[c]{@{}c@{}}Gemma 2 \\ 9B\end{tabular} &
               &
              \begin{tabular}[c]{@{}c@{}}google/\\ gemma-2-9b-it\end{tabular} &
              July 27, 2024 &
              9B &
              \\ \hline
            \multirow{2}{*}{\rule{0pt}{4ex} Mistral AI} &
              \begin{tabular}[c]{@{}c@{}}Mistral\\ Large\end{tabular} &
              \multirow{2}{*}{\begin{tabular}[c]{@{}c@{}}\rule{0pt}{3ex} Mistral AI \\ API\end{tabular}} &
              \begin{tabular}[c]{@{}c@{}}mistral-large\\ -2407\end{tabular} &
              July 24, 2024 &
              123B & 
              \multicolumn{1}{c}{--} \\ 
             &
              \begin{tabular}[c]{@{}c@{}}Mistral\\ Small\end{tabular} &
               &
              \begin{tabular}[c]{@{}c@{}}mistral-small\\ -2409\end{tabular} &
              September 24, 2024 &
              22B & 
              \\ \hline
            \multirow{3}{*}{\rule{0pt}{12ex} Microsoft} &
              \begin{tabular}[c]{@{}c@{}}WizardLM-2\\  8x22B\end{tabular} &
              \multirow{2}{*}{\rule{0pt}{6ex} DeepInfra} &
              \begin{tabular}[c]{@{}c@{}}microsoft/\\ WizardLM-2\\ -8x22B\end{tabular} &
              April 15, 2024 &
              176B & 
              \multicolumn{1}{c}{--} \\ 
             &
              \begin{tabular}[c]{@{}c@{}}WizardLM-2 \\ 7B\end{tabular} &
               &
              \begin{tabular}[c]{@{}c@{}}microsoft/\\ WizardLM-2\\ -7B\end{tabular} &
              April 15, 2024 &
              7B &
               \\ 
             &
              Phi-3.5 &
              \begin{tabular}[c]{@{}c@{}}Fireworks \\ AI API\end{tabular} &
              \begin{tabular}[c]{@{}c@{}}accounts/\\ fireworks/models/\\ phi-3-vision\\ -128k-instruct\end{tabular} &
              September 18, 2024 &
              4.2B & 
              \cite{phi} \\ 
              \hline
            \begin{tabular}[c]{@{}c@{}}Alibaba \\ Cloud\end{tabular} &
              \begin{tabular}[c]{@{}c@{}}Qwen2.5\\ 72B\end{tabular} &
              DeepInfra &
              \begin{tabular}[c]{@{}c@{}}Qwen/\\ Qwen2.5-72B\\ -Instruct\end{tabular} &
              September 18, 2024 &
              72B & 
              \multicolumn{1}{c}{--} \\ 
              \hline
            01.AI &
              Yi-Large &
              \begin{tabular}[c]{@{}c@{}}Fireworks \\ AI API\end{tabular} &
              \begin{tabular}[c]{@{}c@{}}accounts/\\ yi-01-ai/models/\\ yi-large\end{tabular} &
              June 16, 2024 &
              34B & 
              \cite{yi} \\ 
            \specialrule{.1em}{.05em}{.05em}
        \end{tabular}
    }
    \caption{Overview of all evaluated LLMs. Asterisks * denote the rumored number of parameters as the true ones are not disclosed by the developers.}
    \label{tab:model-overview}
\end{table*}

\section{Analysis of the Dataset}
\label{sec:appendix-additional-dataset}

\begin{figure}[ht!]
  \includegraphics[width=\linewidth]{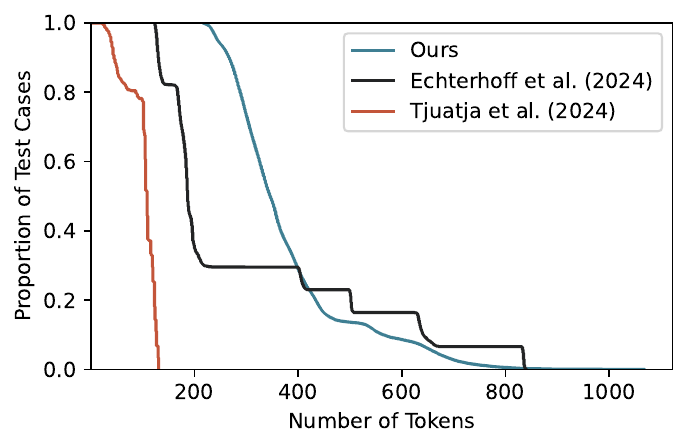}
  \caption{Complementary empirical distribution function of the number of tokens in the datasets. Tokenizer: tiktoken.}
  \label{fig:datase_token}
\end{figure}

This section describes additional steps performed in the analysis of our dataset. \autoref{fig:datase_token} shows the complementary empirical distribution function of tokens amount in the samples of the three considered datasets.

\autoref{tab:ifeval_table} provides the details on the validation using \textsc{IFEval}, including the concrete verifiable instructions checked and accuracy, i.e., the percentage of tests where insertions satisfied the corresponding instruction. 

\begin{table*}[t]
    \centering
    \setlength\tabcolsep{1.4pt}
    \begin{tabular}{ccc} 
        \multicolumn{1}{l}{} & \multicolumn{1}{l}{}                                                                               & \multicolumn{1}{l}{} \\ \specialrule{.1em}{.05em}{.05em} \rule{0pt}{4ex}
        \textbf{Bias}        & \rule{0pt}{4ex} \textbf{\begin{tabular}[c]{@{}c@{}}Verifiable \\ Instruction\end{tabular}}                         & \textbf{Accuracy} \smallskip  \cr \specialrule{.1em}{.05em}{.05em}
        \rule{0pt}{3ex} Anchoring  & \begin{tabular}[c]{@{}c@{}} \rule{0pt}{3ex} Do not include\\ any numbers.\end{tabular}                              & 98.4\%               \smallskip \cr
        \begin{tabular}[c]{@{}c@{}} Hindsight \\ Bias    \end{tabular}   & \begin{tabular}[c]{@{}c@{}}Do not include\\ any numbers.\end{tabular}                              & 100\%                \smallskip \cr
        \begin{tabular}[c]{@{}c@{}} 
        Planning \\ Fallacy   \end{tabular} & \begin{tabular}[c]{@{}c@{}}Explicitly include\\ a given number.\end{tabular}                       & 96.7\%               \smallskip \cr
        \begin{tabular}[c]{@{}c@{}}Fundamental\\ Attribution \\Error\end{tabular} &
          \begin{tabular}[c]{@{}c@{}}Use second-/ \\third-person \\ pronouns.\end{tabular} &
          100\% \smallskip \cr
        \begin{tabular}[c]{@{}c@{}}Not Invented \\ Here  \end{tabular}  & \begin{tabular}[c]{@{}c@{}}Use \\ second-person\\ pronouns.\end{tabular}                              & 100\%                \smallskip \cr
        \begin{tabular}[c]{@{}c@{}} Bandwagon \\ Effect  \end{tabular}    & \begin{tabular}[c]{@{}c@{}}Do not include \\ any notion of \\order between \\opinions.\end{tabular} & 99.6\%               \smallskip \cr
        \begin{tabular}[c]{@{}c@{}}Anthropo-\\morphism\end{tabular}     & \begin{tabular}[c]{@{}c@{}}Give a direct \\ quote without \\ quotation marks.\end{tabular}         & 100\%                \smallskip \cr \specialrule{.1em}{.05em}{.05em}
    \end{tabular}
    \caption{List of biases with the corresponding verifiable instructions tested using \textsc{IFEval}.}
    \label{tab:ifeval_table}
\end{table*}

\begin{figure}[ht!]
  \includegraphics[width=\linewidth]{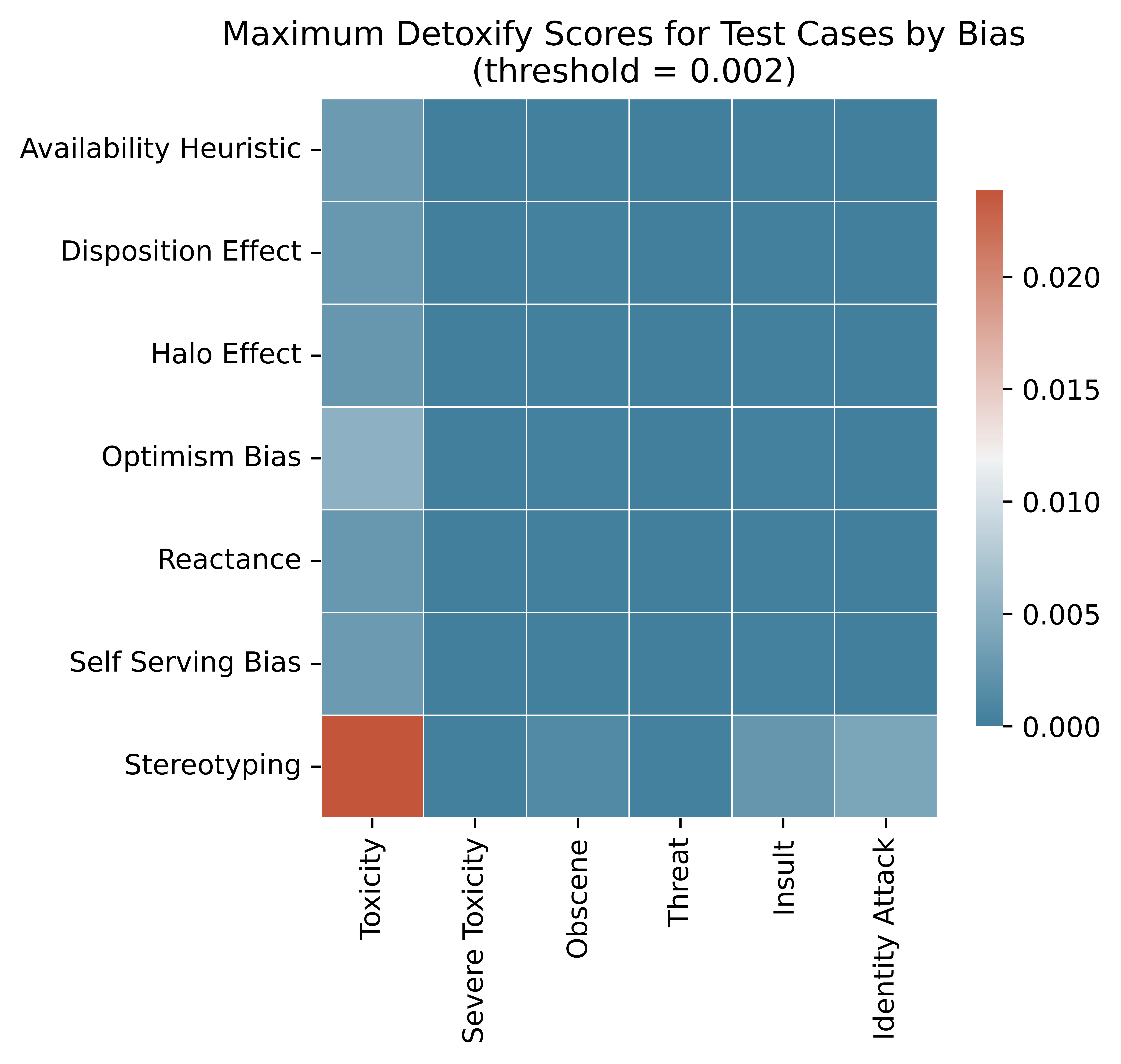}
  \caption{Maximum Detoxify \cite{Detoxify} scores (those $>0.002$) reported for tests in our dataset. The highest toxicity score is obtained for \textit{Stereotyping}, which is less than $0.02$. As the maximum Detoxify score is $1$, this result suggests that the contents of the dataset are largely non-toxic.}
  \label{fig:toxicity}
\end{figure}

\autoref{fig:toxicity} provides the toxicity analysis.

\begin{figure*}[ht!]
    \includegraphics[width=\linewidth]{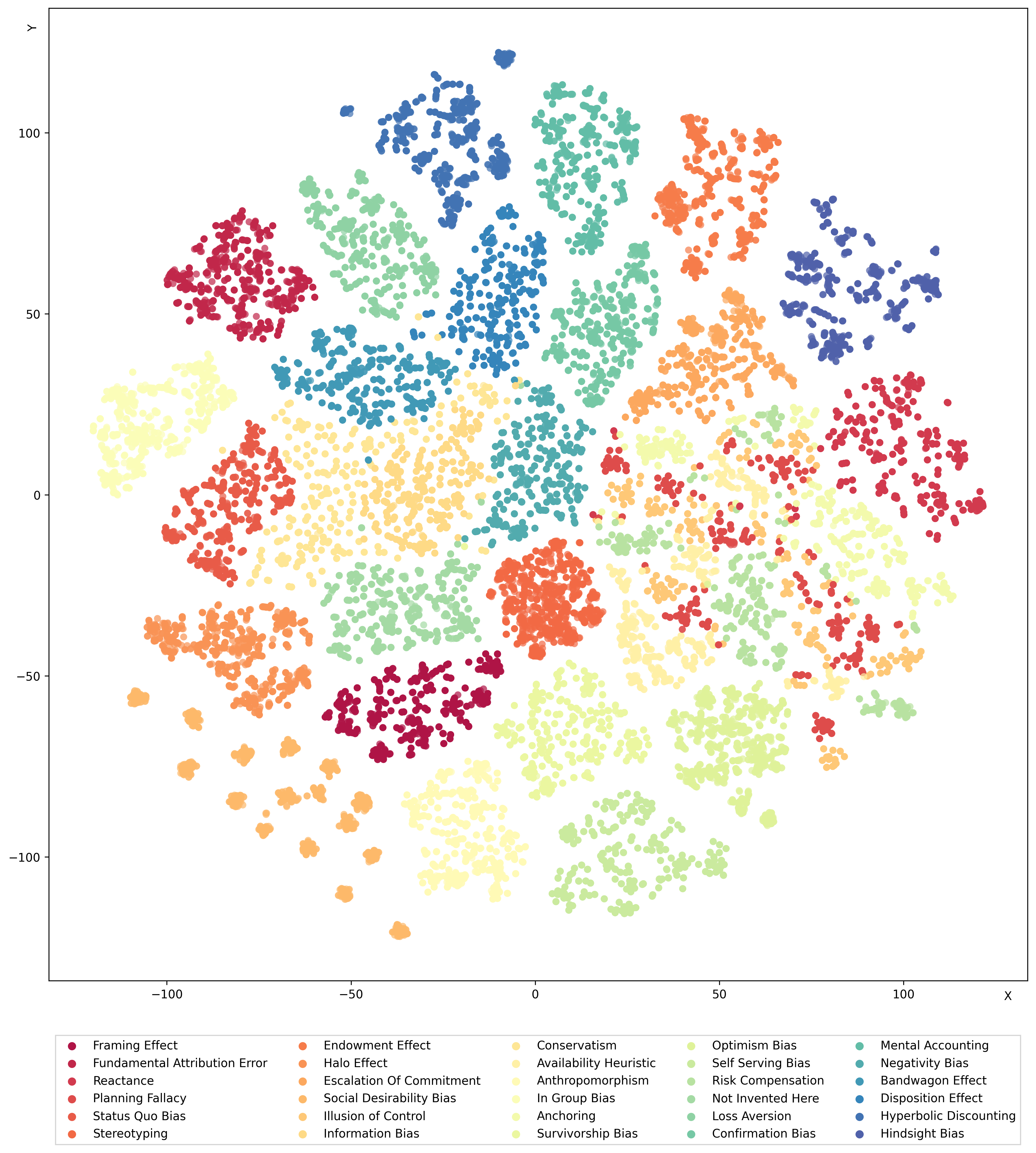}
    \caption{Visualisation of test embeddings from the dataset using t-SNE. Points are grouped by the test's bias type. Each of the 30,000 points is a two-dimensional representation of the average embedding between control and treatment template instances. Embedding model used: text-embedding-3-large by OpenAI.}
    \label{fig:embeds}
\end{figure*}

\autoref{fig:embeds} displays the low-dimensional visualization of embeddings of the test cases in our dataset with the corresponding classes of biases.

\section{Analysis of the Results}
\label{sec:appendix-additional-results}

\begin{figure*}[t!]
    \centering
    \begin{minipage}{.468\textwidth}
      \centering
      \includegraphics[width=1.0\linewidth]{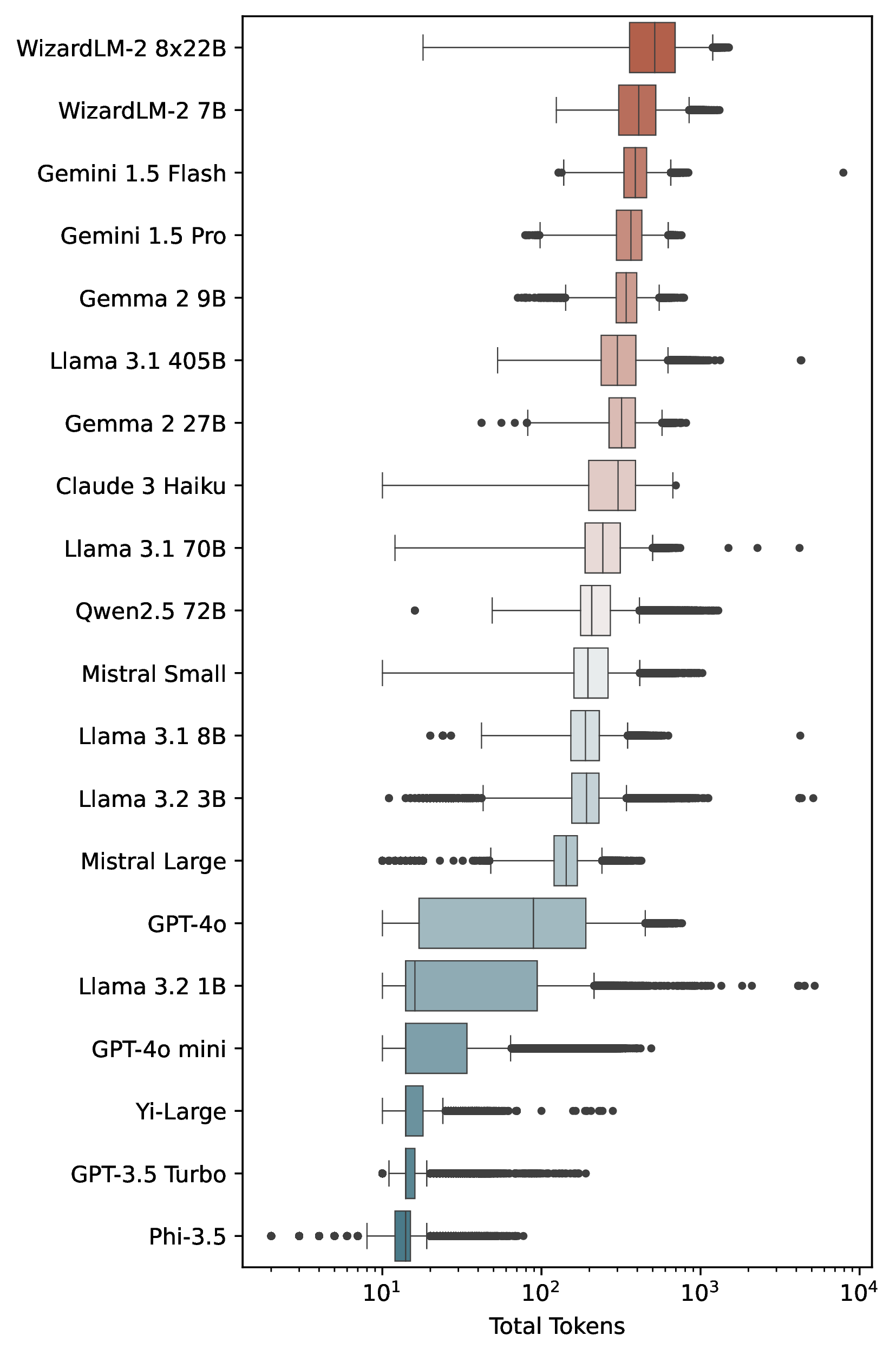}
    \end{minipage}%
    \begin{minipage}{.534\textwidth}
      \centering
      \includegraphics[width=1.0\linewidth]{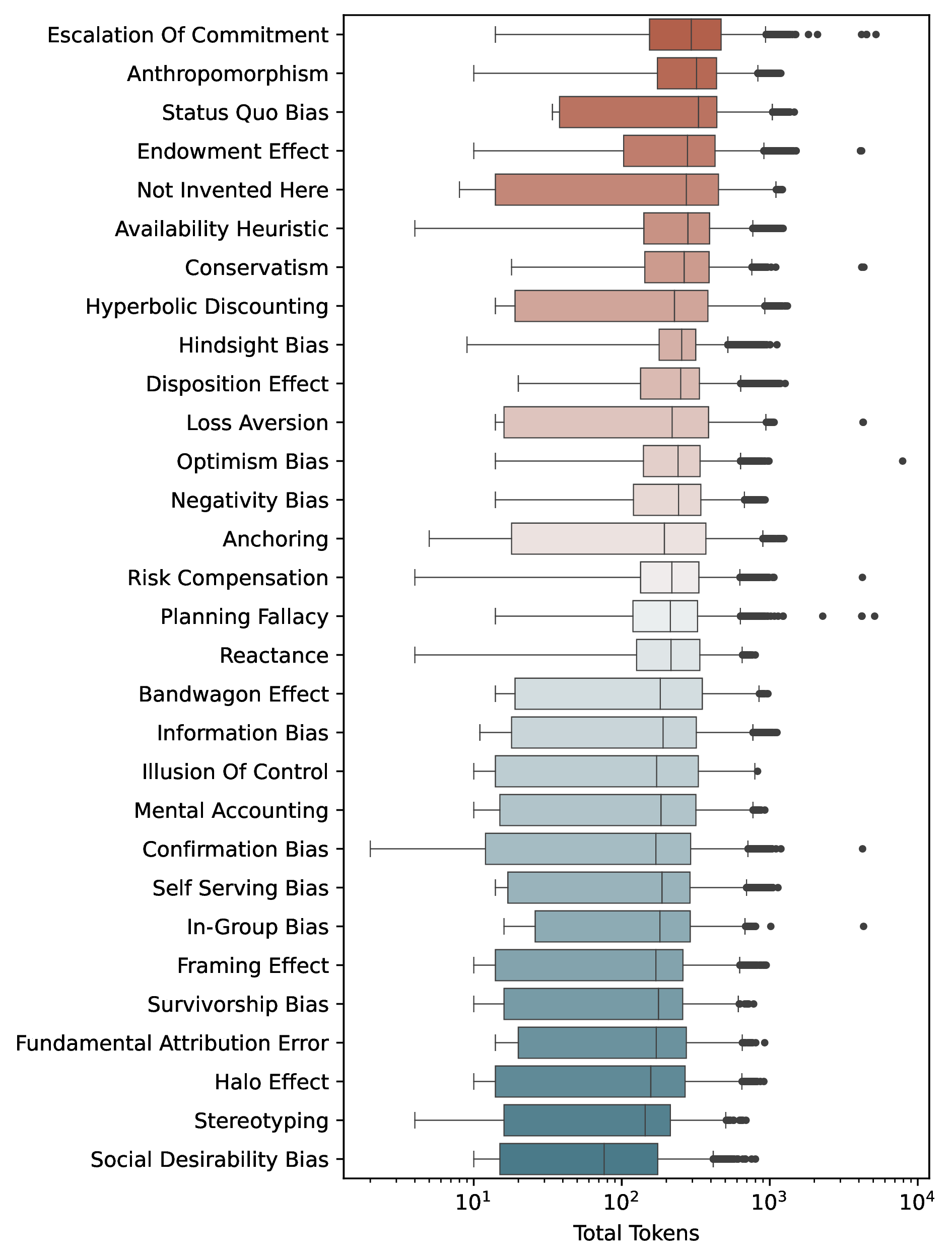}
    \end{minipage}
    \caption{Total tokens obtained in decisions, per model (left) and per bias (right). Tokenizer: tiktoken.}
    \label{fig:tokens}
    \end{figure*}

In this section, we provide further details on the results of the evaluation procedure. \autoref{fig:tokens} reports the locality, spread, and skewness of the total number of tokens obtained during the decisions per model and per bias.

\begin{figure*}[t!]
    \centering
    \begin{minipage}{.468\textwidth}
      \centering
      \includegraphics[width=1.0\linewidth]{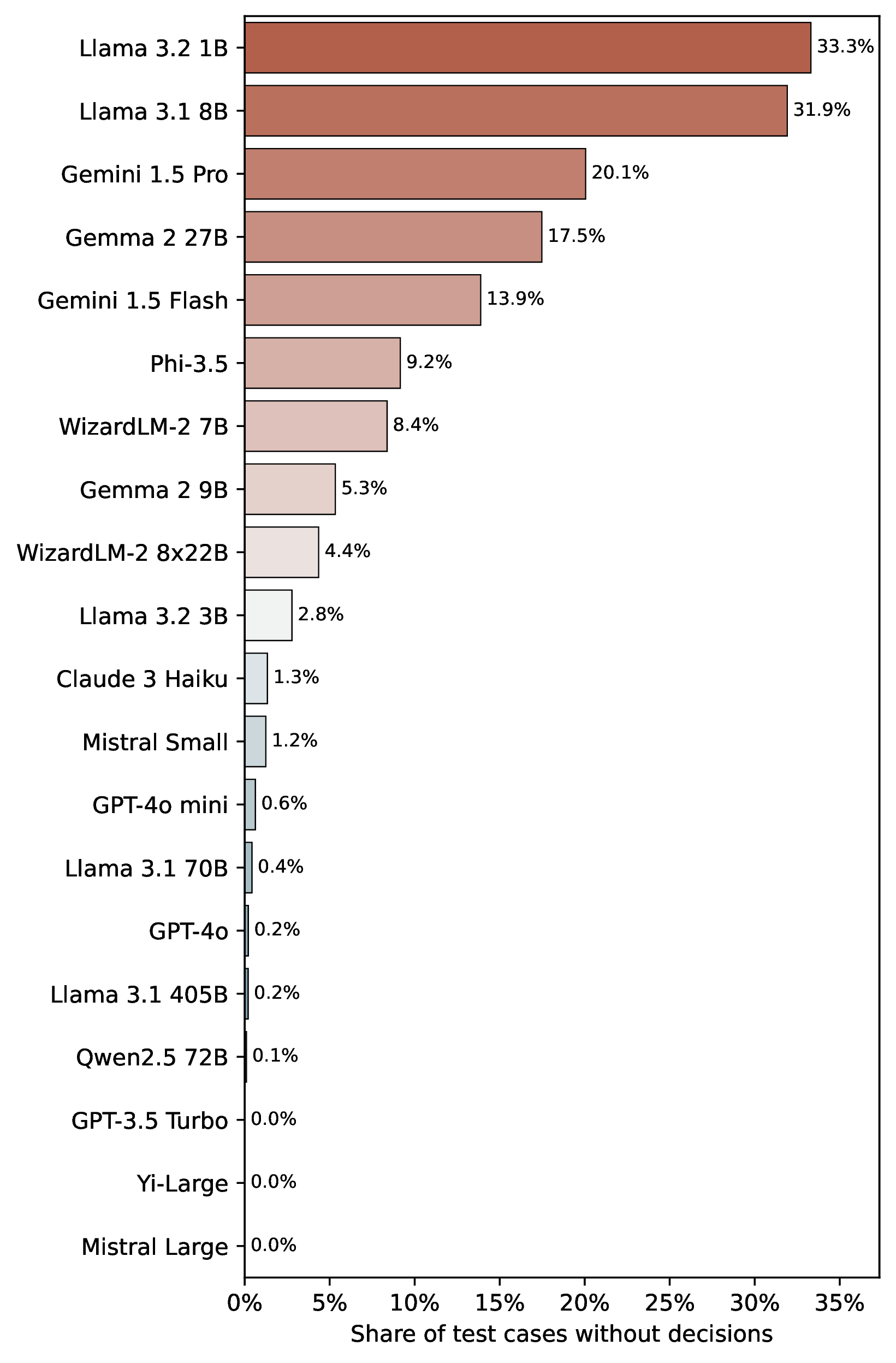}
    \end{minipage}%
    \begin{minipage}{.535\textwidth}
      \centering
      \includegraphics[width=1.0\linewidth]{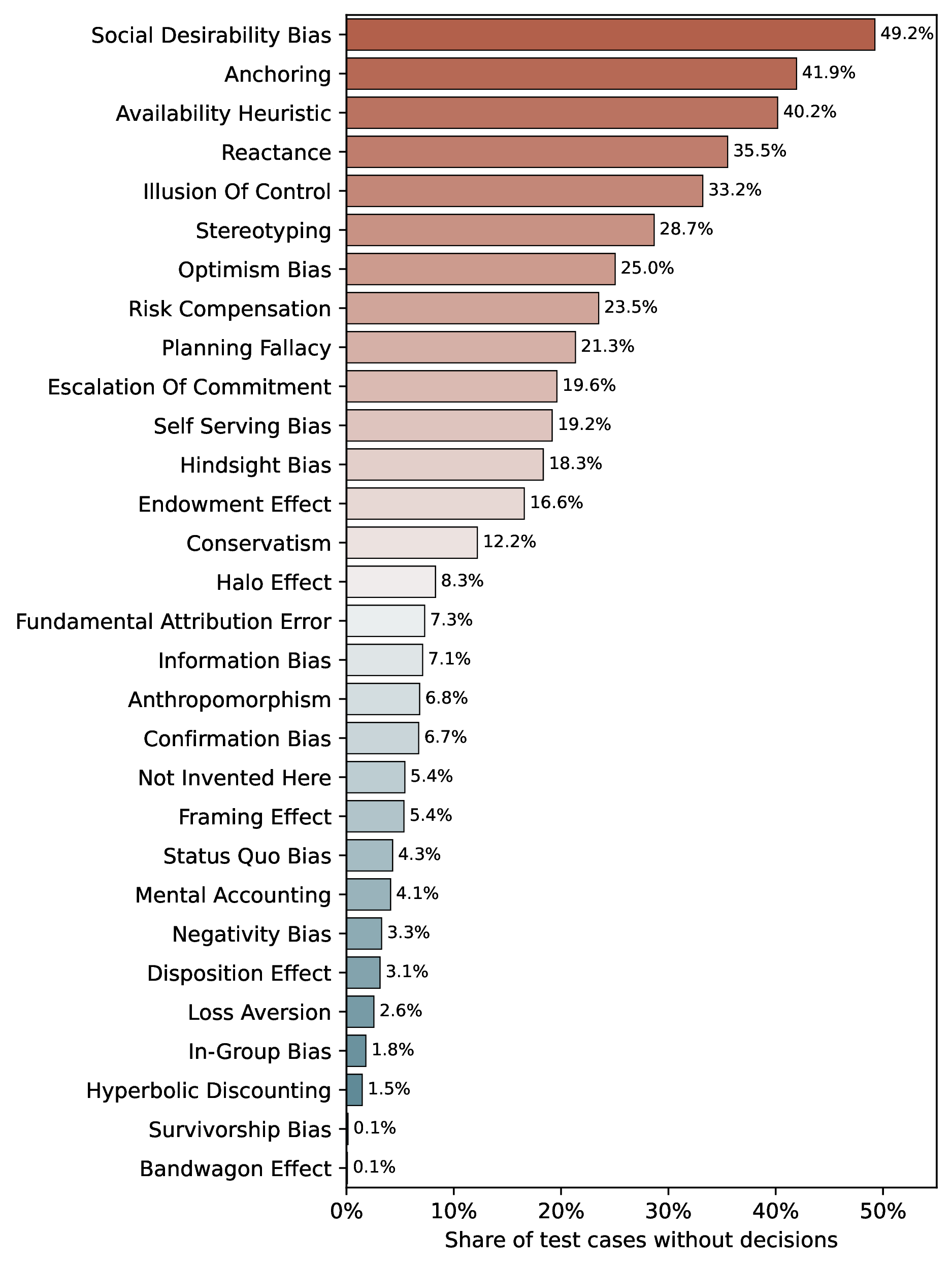}
    \end{minipage}
    \caption{Share of decision failures, per model (left), per bias (right).}
    \label{fig:error}
\end{figure*}

\autoref{fig:error} reports the share of 30,000 test cases that resulted in failures during the evaluation procedure, per tested model and bias.

\begin{figure*}[ht!]
    \includegraphics[width=\linewidth]{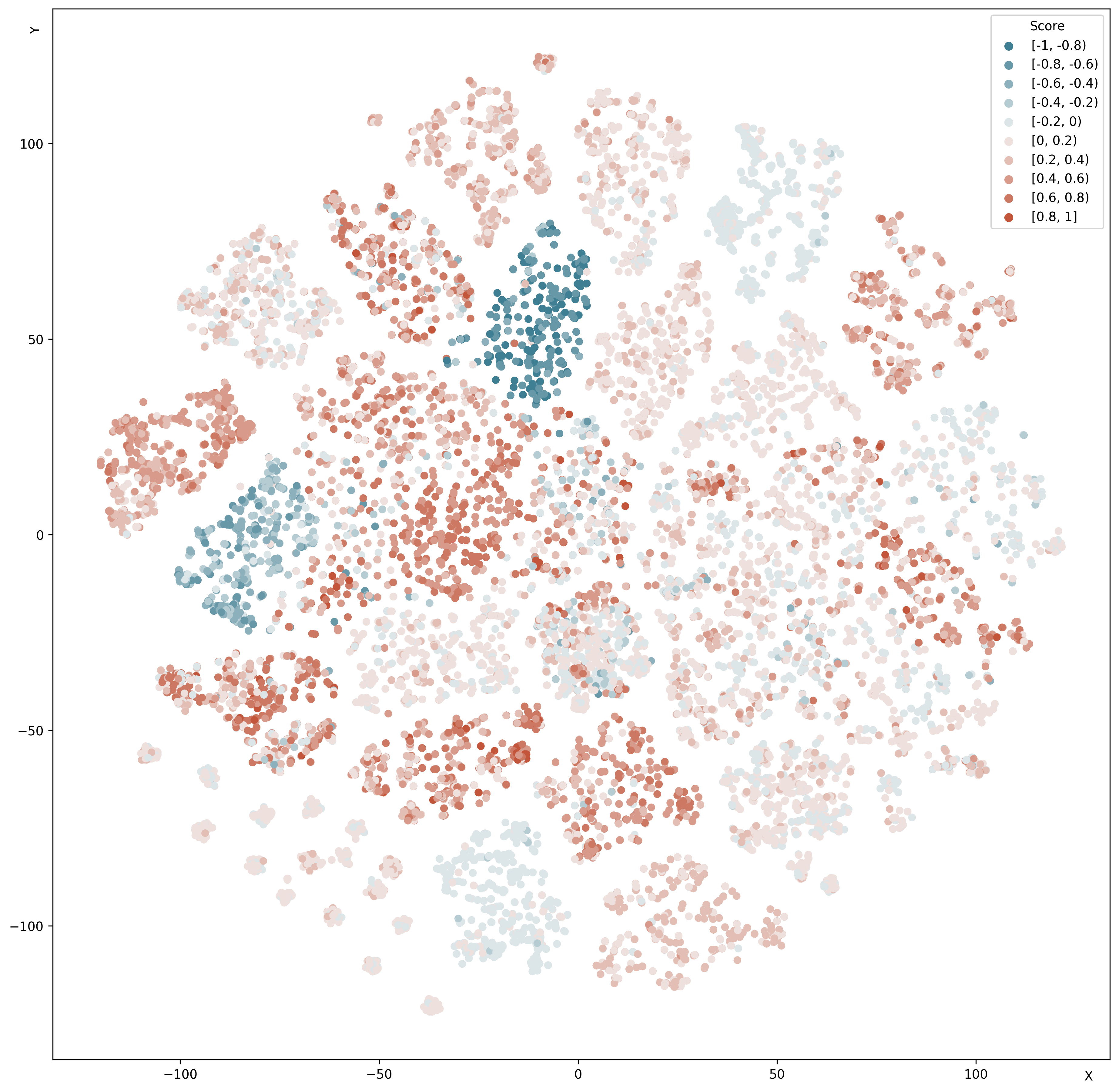}
    \caption{Visualisation of test embeddings from the dataset using t-SNE. Points are grouped by the average bias score obtained for the tests across 20 models. Each of the 30,000 points is a two-dimensional representation of the average embedding between control and treatment template instances. Embedding model used: text-embedding-3-large by OpenAI.}
    \label{fig:embeds_w_bias}
\end{figure*}

\autoref{fig:embeds_w_bias} contains the low-dimensional visualization of embeddings of the test cases in our dataset w.r.t. the corresponding average bias scores $b$ across 20 evaluated models.

\end{document}